\documentclass[preprint,12pt,authoryear]{elsarticle}

\usepackage{amssymb}
\usepackage{amsthm}

\usepackage{hyperref}
\usepackage{booktabs}
\usepackage{multirow}
\usepackage{tikz}
\usepackage[normalem]{ulem}

\journal{Engineering Applications of Artificial Intelligence}

\begin{document}

\begin{frontmatter}



\title{Artificial Intelligence and Natural Language Processing and Understanding in Space: A Methodological Framework and Four ESA Case Studies}


\author[label1]{José Manuel Gómez-Pérez}
\author[label1]{Andrés García-Silva}
\author[label2]{Rosemarie Leone}
\author[label3]{Mirko Albani}
\author[label4]{Moritz Fontaine}
\author[label2]{Charles Poncet}
\author[label2]{Leopold Summerer}
\author[label5]{Alessandro Donati}
\author[label2]{Ilaria Roma}
\author[label5]{Stefano Scaglioni}

\affiliation[label1]{organization={Language Technology Research Lab, Expert.ai},
            addressline={3 Poeta Joan Maragall}, 
            city={Madrid},
            postcode={28020}, 
            country={Spain}}

\affiliation[label2]{organization={European Space Research and Technology Centre (ESA-ESTEC)},
            addressline={Keplerlaan 1}, 
            city={Noordwijk},
            postcode={2201 AZ}, 
            country={The Netherlands}}

\affiliation[label3]{organization={European Space Research Institute (ESA-ESRIN)},
            addressline={Via Galileo Galilei, 1}, 
            city={Frascati},
            postcode={00044}, 
            country={Italy}}

\affiliation[label4]{organization={European Space Agency (ESA)},
            addressline={24 Rue du Général Bertrand CS 30798}, 
            city={Paris},
            postcode={75345}, 
            country={France}}

\affiliation[label5]{organization={European Space Operations Center (ESA-ESOC)},
            addressline={Robert-Bosch-Str. 5}, 
            city={Darmstadt},
            postcode={64293}, 
            country={Germany}}

\begin{abstract}
The European Space Agency is a powerful force for scientific discovery in numerous areas of space
. The amount and depth of the knowledge produced throughout the different missions carried out by ESA and their contribution to scientific progress is enormous and involves large collections of documents like feasibility studies, technical reports, scientific publications, and quality management procedures, among many others. 
Handling such wealth of information, of which large part is unstructured text, is a colossal task that goes beyond human capabilities, hence requiring automation. In this paper, we present a methodological framework based on artificial intelligence and natural language processing to automatically extract information and enable machine understanding of space documents. 
We illustrate such framework through several case studies implemented across different functional areas of ESA, including Mission Design, Quality Assurance, Long-Term Data Preservation and the Open Space Innovation Platform, and demonstrate the value of our approach by solving complex information extraction and language understanding challenges that had not been addressed in space until now. 

\end{abstract}



\begin{keyword}
Space Science and Engineering \sep Artificial Intelligence \sep Natural Language Processing and Understanding \sep Information Management


\end{keyword}

\end{frontmatter}


\section{Introduction}
\label{sec:intro}
The European Space Agency (ESA) is Europe’s gateway to space, with the mission to shape the development of Europe’s space capability and ensure that investment in space continues to deliver benefits to the citizens of Europe and the world. ESA consistently helps to answer the biggest scientific questions of our time, such as the mysteries of the Universe, the understanding of our Solar System, and the quest for life outside our home planet. Its space mission programs are a powerful force for scientific discovery, both looking outward to the confines of our galaxy and beyond to understand the origin of the Universe, as well as inwards, observing Earth through constellations of satellites orbiting our planet to study Earth's climate and define climate change mitigation, adaptation and development pathways. 

The amount, depth and scope of the data, information and knowledge generated and managed during such missions is enormous and their contribution to scientific progress is invaluable. From the announcement of opportunity and feasibility study to space and ground segments design, development, operation, mission decommissioning, and long-term preservation, large collections of heterogeneous information are produced. Some examples include: ideas to develop innovative solutions to technical and operational challenges, space project design and implementation documents, technical reports, operational procedures, quality management instructions, and space records about missions spanning over more than 40 years, like climate data records, exploitation reports, and scientific publications. Managing, mining, and exploiting such wealth of information, of which a large part is free text, is a colossal task that goes beyond human capabilities.

In this paper, we present a methodological framework based on Artificial Intelligence (AI) and Natural Language Processing and Understanding (NLP/U\footnote{Henceforth, we will use NLP indistinctly for both NLP and NLP/U.}) to automatically extract information from text documents related to space missions and enable machine understanding during different mission stages, contributing to create a virtuous circle of knowledge acquisition, management, and transfer at ESA, as well as scientific discovery and innovation worldwide. We demonstrate the added value of this approach through actual NLP solutions implemented at ESA, with potential impact across a wide range of space areas. The goals of such systems range e.g. from assisting the evaluation of the innovation potential of ideas submitted to ESA through the Open Space Innovation Platform (OSIP) to facilitating access to space mission design information, contributing to the adoption of quality assurance and training procedures, and helping to preserve heritage space mission and operation records for long-term archival and exploitation.

This work contributes to unroll the vision of ESA's 2025 Agenda\footnote{\url{https://www.esa.int/About_Us/ESA_Publications/Agenda_2025}} in "adopting fast-learning/higher-risk approach for future technology maturation such as AI” through the implementation of intelligent systems able to support ESA's workforce in several tasks, like effortlessly searching and recommending space information within ESA's repositories, automatically determining how innovative an idea can be, answering questions about spacecraft design or generating training materials to master space operation procedures. The accomplishments described in the paper represent a step forward in increasingly intelligent AI focused on NLP and its applications for information management in space, from assistants able to structure and facilitate access to information to intelligent systems capable to understand and reason with it. We envision a future where AI systems augment human capabilities and engage with human peers in solving challenging technical problems, joining forces in producing major scientific discoveries that could eventually be worthy of a Nobel Prize~\citep{Kitano_2016}.

The remainder of the paper is as follows. Section~\ref{sec:ant} provides an account of the different types of applications of AI and NLP in the scientific enterprise and their relation to space. Next, in section~\ref{sec:fw} we propose our methodological framework for the application of NLP technologies in space. Section~\ref{sec:pick} summarizes the guidelines proposed in section~\ref{sec:fw} and makes special emphasis on the aspects to consider in order to decide whether to adopt a machine learning-based approach to NLP, a symbolic approach or a combination of them depending on the results of the analysis of the use cases and NLP tasks to be addressed. Sections~\ref{sec:spaceqa} to~\ref{sec:osip} illustrate the application of our framework to specific NLP projects recently developed at ESA. Based on such experiences, section~\ref{sec:disc} provides a series of recommendations for the successful development of NLP capabilities in space. Finally, section~\ref{sec:conc} concludes the paper.



\section{Antecedents and related work}
\label{sec:ant}

In her presidential address at the AAAI Conference on Artificial Intelligence,~\cite{Gil_2022} pondered whether AI will write scientific papers in the future. Both her and many others including us believe that we can be hopeful that the answer will be yes and that it may happen sooner than we might expect. Our capabilities to do scientific and technical breakthroughs need to be augmented as scientific questions become significantly more complex. Compare for instance the challenges of formulating Kepler's laws of planetary motion with demonstrating the existence of binary stellar-mass black hole systems~\citep{PhysRevLett.116.061102}. While the former was achieved by one scientist, the latter required a large and interdisciplinary team involving the collaboration of hundreds of scientists from different fields to work together during years to produce results.

\begin{figure}[t]
    \centering
    \includegraphics[width=\textwidth]{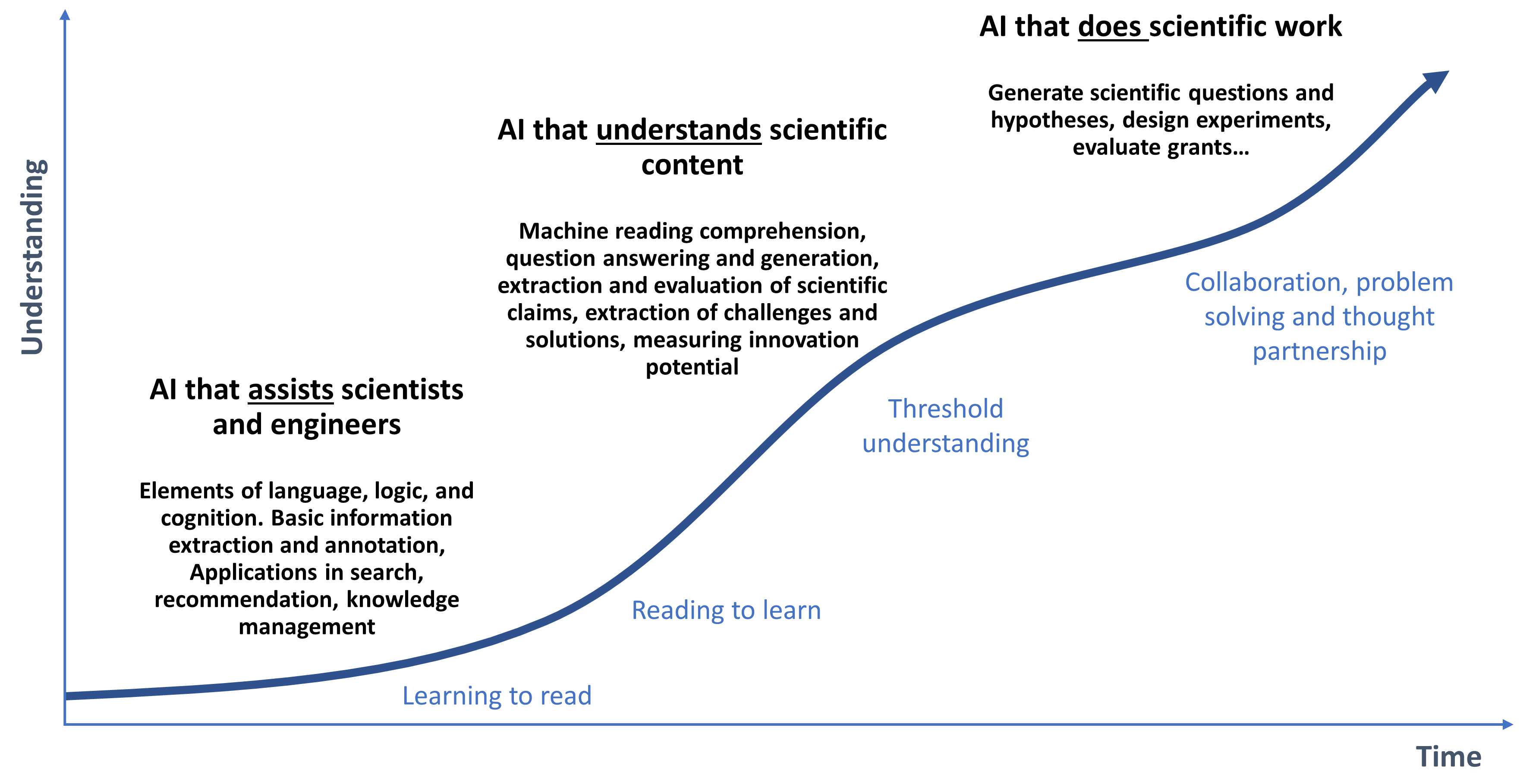}
    \caption{Foreseen progression of the role of AI systems in the scientific enterprise.}
   \label{fig:towardsTAI}
\end{figure} 

Space science and engineering is no exception. The challenges that need to be addressed are extremely complex and involve increasingly large and interdisciplinary teams. AI and specifically NLP become imperative to manage the large volumes of information 
that need to be processed during the lifecycle of space missions. Some examples of scenarios in space where such capabilities are required include the analysis of documents involving e.g. mission objectives definition, mission feasibility and concept analysis, ground segment, space segment and launch segment requirements definition, design and development, satellite platform operations or payload space records acquisition and processing, among many others. In those and other related areas, NLP technologies are starting to prove their value. In some occasions, to extract information from large collections of  scientific documents~\citep{gomez-perez-2017, Murdaca2018KnowledgebasedIE, GARCIASILVA2019550, Berquand2020SpaceMD, BERQUAND2021104273}, producing semantic metadata (see section~\ref{sec:fair}) that enables the development of sophisticated information retrieval applications~\citep{Rico2017CollaborationSA}, making research more accessible in accordance to the principles of FAIR research data~\citep{wilkinson2016fair}, and contributing to long-term data preservation. Other increasingly representative scenarios of application of NLP technologies include systems that address needs related to language understanding of space mission technical documents, like ESA's Concurrent Design Facility (CDF) reports or Quality Management procedures, automatically answering and even formulating questions about space~\citep{garcia-silva-2022a, garcia-etal2022} (sections~\ref{sec:spaceqa} and~\ref{sec:spaceqquiz}).

Like Gil, we  foresee a future scenario (see figure~\ref{fig:towardsTAI}) where AI systems will not only assist but also become an effective part of the scientific and engineering space ecosystem, collaborating, independently pursuing substantial aspects of space mission development,  operation, and space data records analysis, and contributing their own discoveries to the space community. Today, we are already witnessing AI systems that address language understanding challenges involving scientific and technical documents. As originally put by~\cite{Reddy_1988}, "Reading a chapter in a college freshman text and answering the questions at the end of the chapter is a hard (AI) problem that requires advances in vision, language, problem-solving, and learning theory.”. Although this is one of the grand challenges in AI yet to be tackled, recent AI systems like ARISTO~\citep{Clark2019FromT} and ISAAQ~\citep{gomez-perez-ortega-2020-isaaq} are already capable to read a scientific text and answer questions related to its content at a level similar to humans. However, none of such systems have focused on space yet, probably because they rely on key components of modern NLP like transformer language models~\citep{NIPS2017_3f5ee243} that until very recently have not been trained on space data. Such limitation has started to be addressed with the advent of new language models specific for space~\citep{9548078}, which on the other hand still need to prove real-life added value over general-purpose counterparts.

In this paper, we focus on the first two of the three steps of the timeline shown in figure~\ref{fig:towardsTAI}, which represents the evolution of the possible roles to be adopted by AI in the scientific ecosystem, from the perspective of NLP and its applications in space. Therefore, we will delve into two main scenarios. On the one hand, assisting space scientists, engineers and other stakeholders to extract information from text documents. On the other hand, the application of state-of-the-art language technologies to develop AI systems that are able to understand scientific language in the space domain, solving problems involving text comprehension by machines. 

NLP is concerned with the interaction between computers and human (natural) languages, and, in particular, with programming computers to fruitfully process text corpora. Challenges in NLP frequently involve natural language understanding 
with the ultimate goal to connect language with machine perception in tasks such as entity recognition, relation extraction, text classification, and sentiment analysis, as well as others like machine reading comprehension, text generation, conversation, summarization, and translation, to name but a few.

Many applications of NLP focus on text analytics, an interdisciplinary field that also involves computer science techniques from machine learning and information retrieval. The goal of text analytics is to discover novel and interesting information from document collections that is useful for further analysis or strategic decision making. Text analytics tools can extract structured data from unstructured text, classify documents in one or more classes, label documents with categories from taxonomies, and assign a sentiment or emotion to text excerpts, among other functionalities. Such structured information 
is then used to fuel analytic tools and find patterns, trends, and insights to improve tasks such as search, recommendation, knowledge management and, in general, supporting through automation the accomplishment of any task involving text processing. 

Recent breakthroughs in deep learning have made impressive progress in NLP. Neural language models and particularly transformers~\citep{NIPS2017_3f5ee243} like BERT~\citep{devlin-etal-2019-bert} and GPT-3~\citep{brown2020language}, to name some of the most widely-used, are able to infer linguistic and world knowledge from large collections of text that can be then transferred to deal effectively with NLP tasks without requiring too much additional effort. The popularity of machine and deep learning has caused a shift from human-engineered methods to data-driven architectures in text processing, boosted by new and powerful players in the field~\citep{wolf-etal-2020-transformers}. However, despite such popularity, 
there are still gaps that need to be addressed. 

Particularly relevant is the fact that data-driven approaches require large amounts of data to be trained. Neural language models have lessened the requirement of labeled data to address downstream NLP tasks, yet the need for such data has not disappeared. Beyond general-purpose datasets, labeled data is scarce, labor intensive and expensive to generate, and therefore one of the main burdens to leverage data-driven approaches to NLP in business applications. Other issues faced by language models and data-driven approaches to NLP in general include ethical challenges, like gender and racial bias learnt from bias present in the data the models are trained on, as well as a lack of transparency that makes it difficult to explain model predictions and build trust between human users and such models, particularly in domains where regulation demands systems to justify every decision. Furthermore, there is a strong discussion in the NLP community regarding whether data-driven approaches actually have a true understanding of real-world pragmatics and semantics~\citep{bender-koller-2020-climbing}. It is common for generative language models like GPT-3 to produce text that is realistic but also hallucinatory or nonsensical.\footnote{GPT-3 Bloviator: OpenAI’s language generator has no idea what it’s talking about. \\ \url{https://www.technologyreview.com/2020/08/22/1007539/gpt3-openai-language-generator-artificial-intelligence-ai-opinion}} In space, as well as in any scientific discipline, we need models that do not just look thoughtful and able to reason with scientific text, but models that are indeed scientific. 

The improvements in standard benchmark\footnote{For example, the General Language Understanding Evaluation (GLUE) benchmark (\url{https://gluebenchmark.com}) and its updated, more difficult version SuperGLUE (\url{https://super.gluebenchmark.com})} and leaderboard performance in NLP tasks have also come at the cost of increased model complexity and ever-growing amount of computational resources required for training and using current state-of-the-art models. This in turn results into higher entry barriers for organizations that may not have access to large infrastructures, leaving the field to large industrial players, as well as a fast-increasing environmental impact in terms of CO$_2$ emissions. According to a recent survey conducted by~\cite{Michael2022WhatDN}, these are some of the topics that currently raise more concerns amongst researchers in the NLP community. In response to such concerns, several initiatives\footnote{SustaiNLP: Workshop on Simple and Efficient Natural Language Processing\\ \url{https://sites.google.com/view/sustainlp2022}} in the community promote more sustainable NLP research and practices, with two main objectives: to encourage the development of more efficient NLP models and to provide simpler architectures and empirical justification of model complexity.

On the other hand, symbolic, knowledge-based approaches to NLP are generally considered more logically interpretable, explainable, and grounded on semantics and pragmatics. However, they can also suffer from the so-called knowledge acquisition bottleneck. This term, coined by ~\cite{feigenbaum1984}, refers to the difficulty in acquiring knowledge from experts or resources like unstructured text corpora and represent such knowledge in a format that is useful to build intelligent systems. This can occasionally lead to  brittle representations of a domain, systems that are sensitive to corner cases as the amount of data increases, and time-consuming manual modeling activities. In NLP, the knowledge acquisition bottleneck is usually associated with the problem of labeling text corpora at a large scale and in different languages, as reported by~\cite{ijcai2020p687} in the context of multilingual word-sense disambiguation, hindering the creation of both multilingual knowledge bases and manually-curated training sets. 

\section{A text analytics and NLP methodological framework for space}
\label{sec:fw}
The limited availability of annotated datasets for the space domain hinders the application of  classical regimes for model training as well as modern NLP pipelines based on {\it pre-train, fine-tune, and predict}\footnote{Generative language models like GPT-3 propose a new pipeline known as {\it pre-train, prompt, predict}, where a prompt is a piece of text inserted in the input examples so that the task that needs to be solved can be formulated as a language modeling problem in a zero-shot or few-shot training regime.} recently brought about by the advent of neural language models. At the same time, space is a mission-critical business, where errors can result in large economic drawbacks and even the loss of human lives.\footnote{Consider for example the tragedy of the Challenger space shuttle in 1986 or the explosion of the first Ariane 5 flight in 1996.} Therefore, it is key for space AI systems in general and NLP applications in particular to be not only data-efficient, but also explainable~\citep{adadi2018}. 
As illustrated by the commitment of ESA with the fight against climate change,\footnote{ESA Climate Change Initiative \url{https://climate.esa.int}} sustainable, energy-efficient NLP models are also of increasing importance.



\begin{figure}[t]
    \centering
	\includegraphics[width=\textwidth]{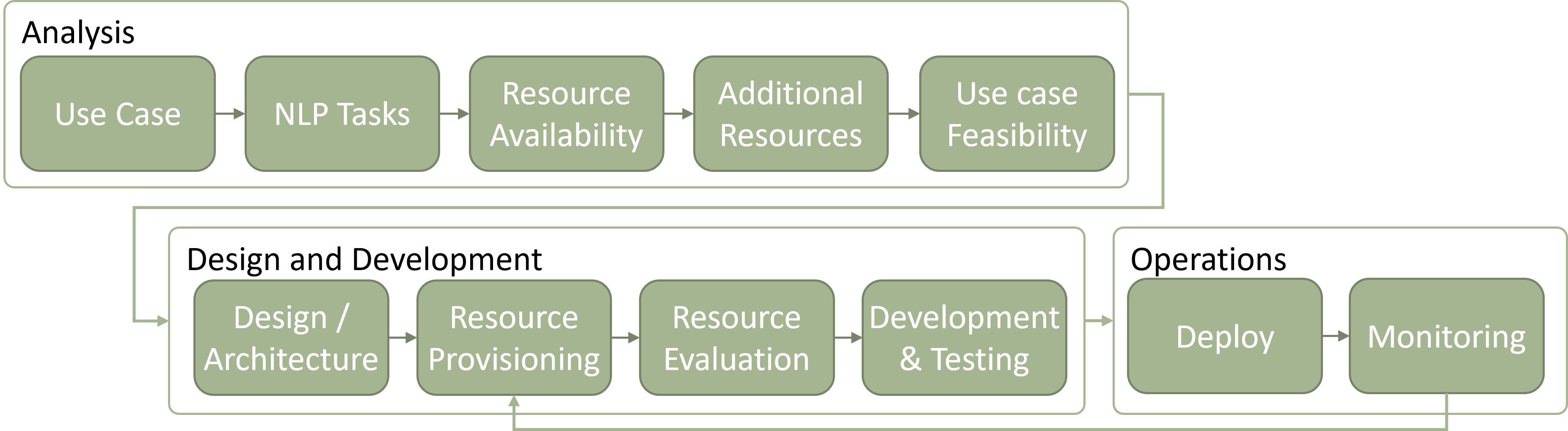}
    \caption{Text analytics and NLP application development workflow in space.}
   \label{fig:wf}
\end{figure}

Figure~\ref{fig:wf} shows a representation of our proposed methodological framework, where we identify three main phases for an NLP project in space: analysis, design and development, and operations. The activities in our framework may appear to be common to modern NLP projects across many domains. However, the idiosyncrasies of NLP projects in space, including those mentioned above (frequent lack of labeled data, limited availability of key NLP components such as domain-specific pre-trained language models, emphasis on explainability, and sustainability), inform the framework itself as well as, in return, its application to address language processing challenges. As a consequence, our framework is particularly aware when it comes to decide whether to adopt a machine-learning based approach, a symbolic approach or a combination of both to develop NLP applications in space.

Many~\citep{Sheth:2017:KPM:3106426.3109448, Shoham:2015:WKR:2859829.2803170, Domingos:2012:FUT:2347736.2347755} argue that symbolic approaches can enhance both expressivity and reasoning power in machine learning architectures and advocate for a hybrid approach that leverages the best of both worlds. For example, in situations where there may be a lack of labeled data, such datasets can be augmented using a knowledge graph to expand the available corpus based on hypernymy, synonymy and other semantic relations represented explicitly in the graph. 
In~\citep{DBLP:books/sp/Gomez-PerezDG20}, we explored the combination of machine learning-based and symbolic approaches in NLP and motivated the application of such paradigm through several success stories in different domains. In this paper, we base on those principles to inform the development of NLP applications in space. However, rather than advocating for a particular approach, we aim at providing the means to make an informed decision in each particular case.

Next, we present the different phases considered by our framework.

\subsection{Analysis}
The analysis phase consists of four main steps focused on the definition of the space-related use cases for NLP technologies, the identification of the NLP tasks that are needed to solve the associated language processing challenges, and the availability of resources to model such tasks. The availability of resources (task-specific annotated datasets, hardware infrastructure, pre-trained models, APIs, structured knowledge) is the most critical step towards discriminating the main approach to address the identified NLP tasks. From a pragmatic point of view, the main principle is usually reusability and cost-benefit ratio. If there are annotated datasets suitable to fine-tune a existing model for the required task in space, probably it is advisable to adopt a machine-learning-based approach. Otherwise, a symbolic approach may be more suitable, e.g., by extending and adapting a pre-existing knowledge graph with terminology extracted from domain text corpora. Next, we describe the steps the analysis phases comprises. Table~\ref{tab:wfex} illustrates the outcome of the analysis phase on several example use cases relevant for ESA.

\begin{table}[ht!]
\footnotesize
  \centering
  \caption{Example use cases of language technologies at ESA, with the analysis of their corresponding NLP task and necessary resources. WSD = Word-Sense Disambiguation, STS = Semantic Textual Similarity, NER = Named Entity Recognition}
  \label{tab:wfex}
  \begin{tabular}{p{2cm}p{4cm}p{1.8cm}p{1.5cm}p{2.5cm}}
    \toprule
    Use case & Description & NLP tasks & General resources & Domain-specific resources\\
    \hline
    \multirow{3}{*}{\parbox{2cm}{Document analytics}} & \multirow{3}{*}{\parbox{4cm}{Extracting key information from space documents like scientific literature or CDF reports, including topics, main concepts, and entities}} & WSD & \multirow{3}{*}{\parbox{1.8cm}{expert.ai NLP API}} & KG extended with space terminology \\
    & & Tax. categorization & & Extended KG, ad-hoc taxonomies \\
    & & NER &  CoNLL-2003 NER & Extended KG\\
    \midrule
    \multirow{3}{*}{\parbox{2cm}{Change Propagation}} & \multirow{3}{*}{\parbox{4cm}{When a change occurs in a document, identify passages in related documents (requirement files, analysis reports, procedures). Similar passages are likely to be affected by this change}} & STS & STS-B & Corpus of space sentences labeled with similarity score 0-5 \\
     & & WSD & Expert.ai NLP API & Space document corpus\\
     & & NER & CoNLL-2003 NER & Space document corpus with annotated entities \\
     \midrule
     Report Quality Checking & Check that CDF reports contain the major elements and set of pre-defined parameters & NER & CoNLL-2003 NER & Corpus of CDF reports with annotated parameter types \\ 
     \midrule
     Automatic Quiz Generation & Given a document, generate the list of the most relevant questions that can be answered with it & Question Generation (NLG) & SQuAD & A corpus of space-related questions and answers marked in the text\\
     \midrule
     Information Retrieval & Retrieve the actual answer to a query from a collection of long documents. Supported queries should include factual questions & Open-Domain Question Answering & SQuAD & A corpus of space-related questions and answers marked in the text\\
  \bottomrule
\end{tabular}
\end{table}


\textbf{Defining the use case.} We start with the definition of the problem that is sought to be addressed through the application of language technologies. Factors to consider include the objectives to be achieved, the type of documents to process, the expected outputs, and how the user is expected to interact with the system. This information can be summarized as a use case name and a description (first two columns in table~\ref{tab:wfex}).
    
\textbf{Mapping the use case against known NLP tasks.} As part of the assessment of the use case, determining its technical feasibility includes the mapping of the use case with the specific set of NLP tasks required to address it. Based on our experience, we identify two main categories of NLP tasks of relevance for space. The first category is related to \textbf{information extraction}, understood as the task of automatically extracting pre-specified types of facts from unstructured documents, producing structured metadata~\citep{Ji2009}. Such metadata may include different types of information, like the main topics or domains related to the text under analysis, the entities that are mentioned in it (people, organizations, locations, time references, quantities, identifiers, etc.), lexical information (surface forms, lemmas), and grammatical information (key phrases, part of speech)
. The output of information extraction tasks is usually connected to information retrieval use cases where the objective is to exploit such metadata to enhance systems like search and recommendation engines, making such systems more expressive and flexible and increasing their coverage. Often, the ultimate goal is to assist users like space scientists and engineers to have easier and more accurate access to relevant information for their daily work. 
    
The other category of NLP tasks in space that we are interested in is related to \textbf{comprehension}. Following~\citep{DBLP:phd/us/Chen18g}, we focus on enabling machines to read some text, process it, and understand its meaning in order to solve tasks like machine reading comprehension, where the system proves its understanding of the text by answering questions about it~\citep{rajpurkar2016squad, choi-etal-2018-quac, kocisky-etal-2018-narrativeqa, reddy-etal-2019-coqa}. In general, we understand as comprehension tasks all those NLP tasks that seek to extract meaning from text in order to solve some language understanding problem. Examples of comprehension tasks would therefore also include: i) summarization, where a system needs to identify the main sentences that provide the most informative summary of a text~\citep{nallapati2017summarunner} or to generate a new, shorter sequence of text that synthesizes the original document~\citep{lin2019abstractive}; ii) question generation, where given a document the system needs to assimilate its content in order to generate plausible questions that can be answered by another system or by a human after reading such text; iii) semantic textual similarity \citep{agirre2013sem}, where the goal is to compare the meaning of two text sequences to determine their relatedness or similarity even if the words and grammatical constructions used for each of them are different; and iv) word-sense disambiguation~\citep{raganato2017word}, which seeks to identify the correct meaning of a polysemic word, e.g. bank, pen or payload, based on its context.
    

\textbf{Assessing resource availability.} Once the NLP tasks relevant for the use case have been identified, the next step involves taking inventory of the language technology resources necessary to address such tasks. Particularly valuable resources include annotated datasets, like SQuAD~\citep{rajpurkar2018know}, that have been labeled and made available by the NLP community to train NLP models in a supervised way to solve downstream tasks like question answering generally, without a specific domain in mind. Other  resources central to modern machine learning-based NLP are pre-trained language models like BERT~\citep{devlin-etal-2019-bert} that were trained in a self-supervised way (no labeling required) on a large collection of general-purpose web documents. Resources useful for NLP tasks also include general-purpose knowledge graphs, e.g. DBpedia~\citep{10.1016/j.websem.2009.07.002} and Wikidata~\citep{10.1145/2629489}, lexico-semantic databases like WordNet~\citep{10.1145/219717.219748}, and  domain-specific document corpora, databases, taxonomies, and thesauri. 
Examples of such resources in space include the Nebula\footnote{Nebula, the knowledge bank of ESA’s R\&D programmes \url{https://nebula.esa.int}} library, which contains a large amount of information in text format about ESA's programmes and studies, the ESA technology tree\footnote{ESA technology tree\url{https://www.esa.int/About_Us/ESA_Publications/STM-277_ESA_Technology_Tree}} or ARTS, ESA's anomaly report tracking system.\footnote{\url{https://artsops.esa.int}} Another type of resources that also needs to be considered but is frequently overseen relates to hardware infrastructure. Machine-learning-based model training and execution need specific GPU infrastructure to work optimally. Also, as mentioned above, training large models on large datasets usually implies a larger carbon footprint compared to a symbolic approach. Therefore, the availability of such resources and whether the project has specific environmental sustainability goals have an  impact on the choice of the most appropriate approach to tackle a given NLP problem.

\textbf{Determining the need for additional resources,} e.g. for model fine-tuning on domain-specific data. As shown in the previous point, most of the available  resources required to solve language tasks are general-purpose and not specialized in the particular domain of interest. Even more so in space, where the availability of domain-specific language resources and particularly annotated datasets for supervised training is  scarce. At the same time, the terminology and in general the language used in space is complex, full of technicalities, and very specific, which hinders the effectiveness of reusing  general-purpose datasets. Therefore, it is often necessary to create domain-specific resources. Among them, we would highlight: i) labeled datasets used to fine-tune a particular machine learning model, usually a pre-trained language model, in order to solve a particular language task in space and ii) the extraction of domain-specific terminologies from a document corpus that can be used to extend the coverage of the domain provided by a general-purpose knowledge graph, if we are opting for a knowledge-based approach. The trade-off between the quality of the predictions obtained and the cost required to create such valuable, domain-specific language resources needs to be taken into account in order to optimize results.

\textbf{Use case feasibility analysis} is possible now that we know the involved NLP tasks, the resources we count with, the resources we need, and the project resources, budget and schedule. If the use case involves an NLP task for which there are not available supporting resources, e.g., domain-specific annotated datasets or GPUs to train deep learning models, then the generation, development and provision of such resources must be considered within the scopes of the project resources, budget and schedule. If provisioning such resources is out of budget due to e.g. personnel, software or hardware cost, the project lacks the necessary skills, e.g., domain experts or knowledge engineers, or time constraints make it impossible to obtain the resources on time, the use case should be discarded. On the other hand, if there are resources supporting the NLP tasks or the generation and provision is feasible we should proceed with the design and development phase.

\subsection{Design and Development}
The design and development phase aims to define the general architecture and software components that make up the solution proposed for the use case, as well as the development of such components. In addition to the traditional software design and development we include activities to generate the required language technology resources, and their evaluation. We also sketch the general components in a high level architecture for NLP-based projects. 

\textbf{Design and Architecture}. As in any software project the goal of the design process is to generate a specification of software components to fulfill the use case or project requirements. Since a general discussion about software design and architecture is out of the scope of this work, we focus on the most prominent components of software project involving NLP components in space. Following a layered architecture pattern \citep{richards2015software}, we define the next layers (bottom-up): 
\begin{itemize}
\item \textbf{Data storage layer}, where text, metadata, and dense representations are stored for efficient retrieval and similarity comparison in an inverted index\footnote{\url{https://lucene.apache.org/}}, an embedding index\footnote{\url{https://github.com/facebookresearch/faiss}} or other document-oriented database.
\item \textbf{Data access layer}, including the components to extract data and text from external sources such as databases, file systems, FTP, web services, and web pages, and components to manage and query the data in the storage layer. Examples of the latter are traditional search engines like Elasticsearch, and neural retrievers such as ColBERT \citep{KhattabColBERT} or DPR \citep{karpukhin2020dense}. In the data access layer, data preprocessing is done including cleaning and formatting text, e.g. text from PDF documents, particularly prevalent in ESA, or web scraping, or generating dense representations for text documents using language models~\citep{reimers-gurevych-2019-sentence}. 
\item \textbf{Domain logic layer} is where software components, including the ones supporting NLP tasks, are orchestrated according to the logic of the different functionalities required for the use case. NLP components include language models fine-tuned for specific task, available NLP APIs or statistical models. An example of an orchestrating component is one that generates questions using a language model, e.g., T5 \citep{raffelT52020}, and then attempts to answer the question using a RoBERTa model \citep{liu2019roberta}. 
\item \textbf{Presentation layer} includes the user interface components necessary for the use case. For example, Kibana dashboards\footnote{\url{https://www.elastic.co/kibana/kibana-dashboard}} are very useful to visualize the results of information extraction tasks on large document collections while enabling search and exploration capabilities. Popular frameworks for rapid app prototyping and demonstration include Streamlit\footnote{\url{https://streamlit.io}} and Dash\footnote{\url{https://dash.plotly.com}}. In addition, we strongly suggest to include feedback gathering mechanism in the user interface. The feedback gathered from users during interaction with the system is a valuable resource to improve the NLP tools and models used in the project, as well as an additional source of supervision for subsequent model training following an active learning approach~\citep{Settles2009ActiveLL}. In integration projects where the visualization of results is in a third party software the presentation layer might contain components exposing the functionalities as web services or API. 
\end{itemize}
  
\textbf{Provisioning of resources} identified in the analysis phase and necessary for the NLP components included in the design. As mentioned above, resources vary widely and include data, corpora, annotation datasets, software, and hardware. From our experience in NLP projects at ESA, it is good practice to invest in the development of at least a testing set to evaluate the performance of the resulting models in the target NLP tasks. In addition, whenever a machine-learning-based approach is adopted, we advocate for the generation of domain-specific annotated datasets for training. To this purpose, we propose to conduct annotation campaigns involving teams that involve experts in space and AI experts in order to guide the former in the text annotation process. We support the annotators with labeling tools like Labelstudio\footnote{\url{https://labelstud.io}} to ease the annotation process. In addition, pre-trained language models are also important resources to be reused in NLP projects, as well as the provision of GPUs to train, fine-tune or run the models.

\textbf{Resource evaluation}
. Training NLP models on poorly annotated datasets leads too underperforming models. Metrics for inter-annotator agreement, like Cohen's~\citep{Cohen1960ACO} or Fleiss~\citep{fleiss1971mns} kappa, are often used to asses the quality of the datasets generated through human annotation. Metrics For NLP models and tools vary depending on the particular NLP task. Common evaluation metrics in NLP include Precision, Recall, F1, Accuracy, BLEU, and ROUGE, to name a few.

\textbf{Development and testing.} With the specification produced in the design activity, the architecture and the selected resources the software development phase can start. As in any other software project, testing is crucial to guarantee the quality of the end product. Agile software development~\citep{shore2021art} has proven useful to deliver software in a timely manner and quickly integrate user feedback in the development process. 

\subsection{Operations}
The final phase is to deploy the software and monitor its performance, actively collecting feedback while providing service to the users. To this purpose, we advocate for modern practices like Dev Ops~\citep{ebert2016devops} and its adaptation to machine learning ML Ops~\citep{sweenor2020mlOps} to enable continuous integration and delivery of the NLP models, bridging the gap between the development, testing, and deployment phases. 

\textbf{Deployment} of the software makes it available for end users. As with any other software, depending on the requirements, we can deploy on-premise, in the cloud, or in the local infrastructure of the technology provider. 

\textbf{Monitoring} the deployed components is important to understand if software is meeting user expectations and use case requirements. Moreover, if feedback mechanisms were included in the user interface, such information needs to be gathered and presented to the development team so that they can use it to improve the NLP models. Furthermore, if the collected data is in the form of annotated data it can be automatically leveraged to re-train the NLP models, following an active learning approach~\citep{Settles2009ActiveLL}. 

\section{Pick your poison: Data hunger vs. the knowledge bottleneck}
\label{sec:pick}

Once the availability of resources has been evaluated, we consider other aspects in addition to performance to choose a machine-learning based approach, a symbolic approach or a combination of them. A priori, resource sufficiency, specially of (annotated) data and GPU infrastructure, tends to advise for a machine learning-based approach. On the other hand, strong requirements in terms of factors like explainability, freedom from bias, and carbon footprint may advise to follow the symbolic route. Nevertheless, recent work in explainable machine learning~\citep{lime,shap,ribeiro-etal-2020-beyond} is pushing the envelope to make such models more transparent to humans and therefore easier to explain and inspect
. Finally, we may also opt for a hybrid approach, e.g. to inject structured knowledge about entities and the relations between them from a pre-existing knowledge base into a pre-trained language model~\citep{peters-etal-2019-knowledge, wang-etal-2021-k}, enabling domain adaptation at a limited cost. However, the latter still needs further research for its systematic application in a production environment.

Many NLP tasks can be solved equally well by using a machine learning-based approach or a symbolic approach. The decision needs to be informed with different criteria according to our framework. First, we look for existing tools or models in the state of the art for each NLP task. If more than one tool or model exists, then some evaluation needs to be carried out to choose between the different options. This is often the case for information extraction tasks of common key information like keywords, phrases, and entities. Nevertheless, when information extraction is performed on ad-hoc information, e.g. specific entity types corresponding to spacecraft instruments or satellite launchers, it is unlikely to find tools or datasets available for reuse. In this case, it is advisable to either extend a pre-existing knowledge base, e.g. a knowledge graph, with domain terminology and write rules following the symbolic approach, or to annotate domain-specific text with labels corresponding to the types of information we need to extract and then train a machine learning model to do the extraction. 

Either way, gathering a representative document corpus of relevant text is often necessary. In the former case the text is analyzed by knowledge engineers with the help of domain experts to elicit the domain terminology to be included into the knowledge graph and write inference rules. Alternatively, in the latter scenario domain experts shall label enough text so that a data scientist can train and evaluate a machine learning model from scratch or (preferably) fine-tune a pre-trained model on the resulting dataset. 
In general, and also in space~\citep{gomezperezaistar2021}, both options may involve considerable effort and have their pros and cons.

When the project requires other capabilities rather than information extraction, involving e.g. comprehension tasks such as closed-book~\citep{Wang2021CanGP} and open domain~\citep{yang-etal-2019-end-end} question answering or text generation~\citep{radford2019better}, the knowledge-based approach might be less appealing due to the potentially large size and complexity of the rule base and knowledge representation that would be required to address the task. This is an instance of the knowledge acquisition bottleneck where, due to the cognitive limitations that a knowledge engineer may experience to identify, formulate in their mind, and explicitly represent the potentially vast number of possible cases to be covered, the resulting model may have difficulties to generalize and suffer from low recall. In this case, machine learning-based or hybrid approaches tend to be more promising. Machine learning models and more specifically neural networks pre-trained for language modeling, have shown good results in such tasks. However, note that even though nowadays we can find generalist resources for comprehension tasks like machine reading comprehension (SQuAD~\citep{rajpurkar2018know}, Natural Questions~\citep{kwiatkowski-etal-2019-natural}, TriviaQA~\citep{joshi2017triviaqa}), such datasets might not be optimal for space. The mismatch between the datasets and the use case could be not only at the terminology level, but also in the type of questions and expected answers. For example, several datasets only support factoid questions where answers are short facts. If the use case requires answer types other than facts, such as explanations, comparisons, list of items or process descriptions, existing datasets are of little help. Something similar occurs in summarization, where available datasets like CNN/Daily Mail~\citep{nallapati-etal-2016-abstractive}, Gigaword~\citep{rush-etal-2015-neural}, and X-Sum~\citep{xsum-emnlp} are mostly centered on news. In such case, it is necessary to create an annotated dataset, 

Machine learning aims to construct algorithms that are able to learn to predict a certain target output. To achieve this, the learning algorithm is presented some training examples that demonstrate the intended relation of input and output values. Then the learner is supposed to approximate the correct output, even for examples that have not been shown during training. Such inductive bias~\citep{Mitchell80theneed} is created by adding statistically significant amounts of examples during model training. Therefore, when machine learning models provide erroneous predictions, the alternatives to fix such errors are usually limited to training or fine-tuning the model in a new setup, i.e. with new hyperparameters, loss function or a slight modification in the architecture, or to provide more annotated data in the hope that the error will be fixed once the model is re-trained. To generate more annotated data we advocate for providing a dedicated GUI that enables gathering feedback from users in the form of additional supervision, e.g. by verifying or refuting the predictions made by the model, or directly providing the correct prediction. However, neither training in a new setup nor using more annotated data is a guarantee to solve erroneous predictions. In such case, a possible solution is to use post-processing rules on the model output and fix the recurrent errors. 

In the following sections we illustrate the application of our methodological framework to four case studies in space. We will show how we addressed the language processing needs that such case studies entail and the decisions we made to successfully accomplish them. Table~\ref{tab:ucdim} characterizes the different case studies in terms of a selection of the key aspects discussed in this section that are particularly relevant for them.

\begin{table}[ht!]
\footnotesize
  \centering
  \caption{Use cases vs. decision criteria. IE = Information Extraction, Comp = Comprehension, XAI = Explainable, Ann. data = labeled data availability, ML/KB =  chosen approach. \checkmark* = annotated datasets exist but not for the domain of interest.}
  \label{tab:ucdim}
  \begin{tabular}{p{5cm}cccccc}
    \toprule
    Use case & IE & Comp. & XAI & Ann. data & ML & KB\\
    \hline
    Answering questions about the design of space missions and spacecraft concepts & x & \checkmark  & \checkmark & \checkmark* & \checkmark & x \\
    \midrule
    Generating quizzes to support training on quality management and assurance in space science and engineering & x & \checkmark & \checkmark & \checkmark* & \checkmark & x\\
    \midrule
    Information extraction for Long-Term Data Preservation in space & \checkmark & x & x & x & x & \checkmark \\
    \midrule
    Assisted evaluation of the innovation potential of OSIP ideas & \checkmark  & \checkmark & \checkmark & x & x & \checkmark\\
  \bottomrule
\end{tabular}
\end{table}

\section{Answering questions about the design of space missions and spacecraft concepts}
\label{sec:spaceqa}

The definition and assessment of future space missions or spacecraft concepts at ESA is undertaken by a multidisciplinary group of experts at the Concurrent Design Facility (CDF), where concurrent engineering principles are applied to speed up design while ensuring consistent and high quality results~\cite{Bandecchi1999ConcurrentEA}. The CDF produces studies that establish the technical, programmatic, and economic feasibility of ESA's endeavours ahead of industrial development. Since its inception in 1998, the CDF has performed more than 250 studies with their associated reports. Typically, a CDF report\footnote{Public CDF reports available at: \url{https://www.esa.int/Enabling_Support/Space_Engineering_Technology/CDF/CDF_Reports}} is a long (200 to 300 pages) document in English that covers a large variety of technical topics related to the mission itself, the systems embedded in it, their configuration, payload, service module, ground segment and operations, and technical risk assessment, to name some of the most common ones. Finding specific pieces of information in such long and complex documents using traditional search engines is a cumbersome and prone-to-error task. In it, a keyword-based query is typically issued to retrieve documents where the information being sought may be contained. Since the exact location of the answer is unknown, then such documents need to be manually inspected.

\begin{figure}[t]
    \centering
    \includegraphics[width=0.8\textwidth]{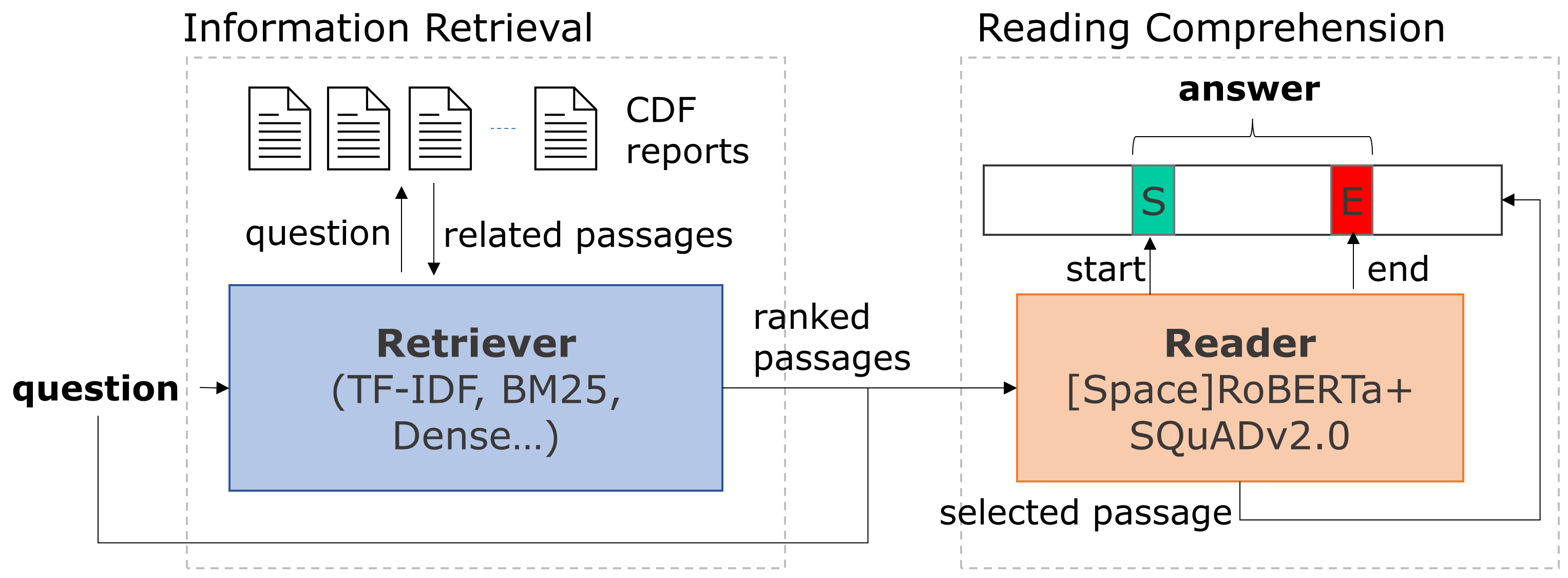}
    \caption{High-level architecture and main components of SpaceQA.}
   \label{fig:spaceqaarch}
\end{figure}

To address these shortcomings, we formulate the problem of retrieving answers to space questions from CDF reports as an open-domain question answering (QA) task~\cite{prager2006open}, which aims to answer a natural language question against a collection of large text documents. Open-domain QA has been a longstanding problem in NLP, information retrieval (IR) and related fields~\citep{chen-yih-2020-open, Voorhees99thetrec-8,moldovan-etal-2000-structure,brill-etal-2002-analysis,Ferrucci_Brown_Chu-Carroll_Fan_Gondek_Kalyanpur_Lally_Murdock_Nyberg_Prager_Schlaefer_Welty_2010}. However, based on recent advances in neural reading comprehension, open-domain QA systems have experimented an accelerated evolution. Complex pipelines involving many different components such as question processing, document and passage retrieval, and answer processing, have been replaced with modern approaches that combine IR and neural reading comprehension~\citep{chen-etal-2017-reading,yang-etal-2019-end-end,DBLP:conf/emnlp/MinCHZ19}. 

This case study falls in the category of comprehension-related problems introduced in section~\ref{sec:fw}. As summarized in table~\ref{tab:ucdim}, some ability to justify why a particular answer is proposed for a given question is required. As we will see, this is addressed by providing the user with a combination of the confidence score about a particular answer produced by the question answering model and the visualization of the actual document context where the answer appears, contributing to the plausibility of the answer. In this case, we do count with a labeled dataset for the question answering task. However, such dataset is not domain-specific. Similarly, the state of the art in NLP also provides us with (general-purpose) pre-trained language models that can be fine-tuned over such data. Finally, to model this problem following a knowledge-based approach, e.g. as a rule-based production system, we would need to anticipate all the questions that a potential user could pose to the system, which is not feasible. Given the combination of all these factors, we opt for a machine learning-based approach with transformer language models at its core.

SpaceQA~\citep{garcia-silva-2022a}, the main result of this activity, is the first implementation of an open-domain QA system for space mission design. As the first system of its kind, SpaceQA needs to face important challenges and limitations. We are aware that space documents and especially CDF reports use complex and domain-specific terminology~\citep{Berquand2020SpaceMD}. 
However, the most limiting factor for the development of SpaceQA is the scarcity of domain-specific resources that can be used in combination with state-of-the-art NLP architectures. In spite of promising recent work in transformer-based language models for space science and engineering~\citep{9548078}, the absolute lack of annotated data for open-domain QA in space prevents a strategy based on fine-tuning. Therefore, a transfer learning approach was adopted that leverages pre-trained language models fine-tuned on available general-purpose datasets for similar tasks.  

\subsection{Approach}
\label{subcsec:spaceqaapp}
The goal of SpaceQA is to find the answer to a factual question about space mission design as a text span from a collection of CDF reports. Since answers need to be extracted from passages from a set of documents, we can catalogue SpaceQA in the category of extractive~\citep{rajpurkar2018know} and open-domain~\citep{yang-etal-2019-end-end} QA systems. 

As shown in figure~\ref{fig:spaceqaarch}, SpaceQA follows a two-stage retriever-reader architecture consisting of: i) a passage retriever component that finds the passages that may contain an answer to the question from a collection of CDF reports and ii) a neural reader component that extracts the answer from some of such candidate passages. For the retriever we evaluate  different methods including traditional sparse vector space methods based on TF-IDF, BM25 or cosine similarity, as well as dense representations using bi-encoders, like \textbf{Dense Passage Retrieval (DPR)}~\cite{karpukhin2020dense}, \textbf{ColBERT}~\cite{KhattabColBERT}, and \textbf{CoCondenser}~\cite{gao-callan-2021-condenser}. The reader is based on state-of-the-art reading comprehension models built on modern transformer architectures. Once a reduced set of the top-k potential passages have been identified by the retriever, the reader attempts to spot the answer to the question as text spans from any of the passages, assigning a score to each of the extracted candidate spans, ranking the set of potential answers. Due to the lack of a question answering dataset for space to train the reader, we resort to the widely used Stanford Question Answering dataset (SQuAD2.0) proposed by~\cite{rajpurkar2018know}. Our first candidate reader model is based on a \textbf{RoBERTa}~\citep{liu2019roberta} language model fine-tuned on SQuAD2.0. Since RoBERTa was pre-trained on a general-purpose corpus, there could be vocabulary mismatch between the set of questions about space mission design in our evaluation dataset and the knowledge encoded in the language model that could affect performance. To bridge that potential gap, we also evaluate \textbf{SpaceRoBERTa}~\citep{9548078}, a version of RoBERTa pre-trained on documents from space science and engineering. SpaceRoBERTa model started from a RoBERTa base model that was further trained on a 14.3 GB corpus of publications abstracts, books, and Wikipedia pages related to space systems.

\subsection{Evaluation}
\label{subsec:spaceqaeval}
We use a manually crafted dataset of factual questions produced by ESA, with answers and paragraphs extracted from CDF reports where the answer to the question has been annotated by an expert. Such test set contains 60 questions, answers, and corresponding paragraphs. While small, this dataset is still useful to evaluate the reading comprehension model by testing whether the right answer is extracted for a question. However, this dataset is particularly limited when it comes to evaluate the passage retrieval module since potentially more than one paragraph can contain the answer for a given question. Thus, we manually search for additional paragraphs containing the answer to the question and extend the dataset with up to 5 more paragraphs for each question. 

\begin{table}[ht!]
  \centering
  \caption{Evaluation of the candidate retriever components.}
  \label{tab:spaceqaretrievers}
  \begin{tabular}{cccc}
    \toprule
    Retriever & R@10 & MRR@10 & Accuracy\\
    \midrule
    TF-IDF & 0.252 & 0.395 & 0.483  \\ 
    BM25 & 0.326 & 0.254 & 0.55  \\ \hline
    DPR & 0.218  & 0.170 & 0.35 \\
    ColBERT & \textbf{0.4898}  & \textbf{0.560} & \textbf{0.717} \\
    CoCondenser & 0.354 & 0.404 & 0.583  \\
  \bottomrule
\end{tabular}
\end{table}

\begin{table}[ht!]
  \centering
  \caption{Evaluation of the reader candidates.}
  \label{tab:spaceqareaders}
  \begin{tabular}{cccc}
    \toprule
    Model & Precision & Recall & F-Score \\
    \midrule
    RoBERTa base & 0.774 & \textbf{0,751} & \textbf{0.762} \\
    RoBERTa large & 0.629  & 0.664 & 0.646 \\ 
    SpaceRoBERTa & \textbf{0.816} & 0.671 & 0.737 \\
  \bottomrule
\end{tabular}
\end{table}

Since the area of open-domain QA in space mission design is largely unexplored, and hence there is a lack of systems to compare against, we focus our evaluation on acquiring an understanding of the current limitations of SpaceQA. To this purpose, rather than overall end-to-end performance we are particularly interested in each of the steps involved in SpaceQA individually. Table~\ref{tab:spaceqaretrievers} shows the evaluation results of the retrievers using recall and mean reciprocal rank (MRR) at 10. We also measure the accuracy of the retrievers to find within the top-10 at least one passage containing the answer. The results obtained in our test set seem to be consistent with those reported in the original papers (except DPR), supporting in this case our decision to adopt a transfer learning approach for passage retrieval. In all three metrics, ColBERT was the best retriever, followed by CoCondenser. Training on MS MARCO~\citep{nguyen2016ms} seems to be key when reusing such retrievers. Trained on Natural Questions~\citep{kwiatkowski-etal-2019-natural}, DPR performs last in our test set. 

Regarding the readers, it has been reported that for natural language questions there is often a number of acceptable answers as well as a genuine ambiguity in whether an answer is acceptable~\cite{kwiatkowski-etal-2019-natural}. For example, for the question \textit{why is the rover top part larger than the bottom part?} an acceptable answer is \textit{to support the solar panels} but \textit{to support the solar panels stack namely 910mm by 500mm} is acceptable, too. A span exact match evaluation could result on low and discouraging results, which would not reflect the actual performance of the reader. Thus, we opt for a token-based evaluation of the reader where we compare the tokens in the span extracted by the reader against the lists of tokens in the ground truth answer. Evaluation results are presented in table~\ref{tab:spaceqareaders}. Unlike for the retrievers, these results illustrate the performance impact of the lack of domain-specific data for fine-tuning, suggesting a considerable room for improvement in that direction. As a reference, the current SotA in extractive QA over SQuAD2.0,\footnote{\url{https://rajpurkar.github.io/SQuAD-explorer}} is currently above 93.2 F1. Interestingly, the reader resulting from fine-tuning RoBERTa base on SQuAD produced the best results, followed by a model based on the domain-specific SpaceRoBERTa also trained on SQuAD. The precision obtained by the SpaceRoBERTa model is higher than with RoBERTa base. However, its recall is lower.


\subsection{The SpaceQA system}
\label{subsec:spaceqasys}
We build an open-domain QA system using the retriever and reader that performed best in our evaluation: ColBERT as retriever and RoBERTa-base fine-tuned on SQuAD2.0 as reader. 
We index the text from the passages extracted from the CDF reports in Elasticsearch. Then, we use FAISS~\citep{JohnsonFAISS}, an efficient library for similarity search and clustering of dense vectors, to index the passage encodings generated with ColBERT. 

When an user asks a question through the system's frontend,  SpaceQA uses ColBERT to encode the question and retrieve from FAISS the 10 most relevant passage ids. Then the passage ids are used to retrieve from ElasticSearch the text of the passages. Next, the question and relevant passages pairs are processed by the reader and the extracted answers are returned along with a confidence score. However, only answers with a score above an empirically defined threshold of 0.5 are shown directly to the user. To see answers with a lower score, the user is prompted first with a warning message about the low certainty of the answers to be displayed. 

\begin{figure}[h!]
  \centering
  \includegraphics[width=0.75\columnwidth]{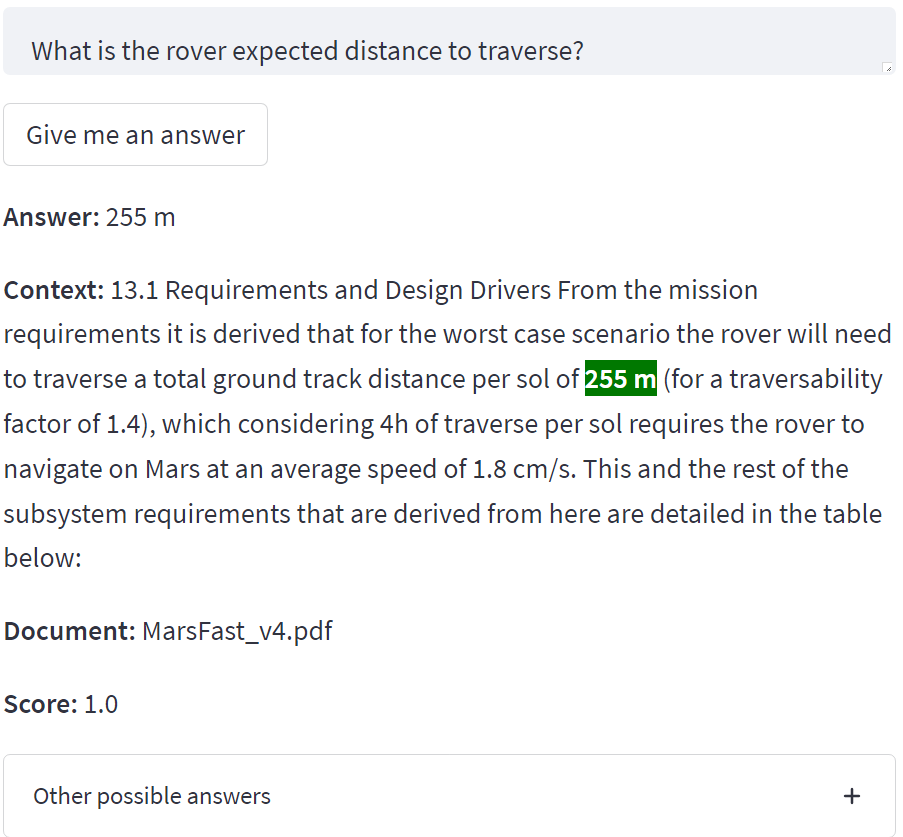}
  \caption{The SpaceQA web application. A screenshot of the user interface.}
  \label{fig:spaceqascreenshot}
\end{figure}

We deployed the system in a machine with 32BG RAM, 1TB SSD, intel i7 CPU, and NVIDIA GeForce 1080Ti GPU. Both the reader and FAISS run on the GPU while the rest of the components use the CPU. Figure \ref{fig:spaceqascreenshot} shows a screenshot of the SpaceQA web application. When the user makes a question, the system displays the answer with the highest score and the passage it was extracted from. To support the plausibility of the answer produced by the model, we highlight the text span corresponding to it in the passage where it was extracted from, providing the user with its context. We also display and link the source document of the passage and the answer score, obtained by multiplying the probabilities of the span start and end tokens generated by the reader. If SpaceQA identifies other possible answers, they are displayed ranked by score under the option "other possible answers". When the answer score is below 0.5, the user is required to actively click on the option to see it. If the reader does not find any answer, we also inform the user.

Since most of the users are used to keyword-based queries in traditional search engines, we also provide a list of predefined questions so that they can experiment with the system features and see examples of the type of questions the system handles. This helps addressing the \textit{blank-page syndrome}.\footnote{The blank page syndrome usually refers to writer’s block, which describes the creative blocks and avoidant thought patterns that many writers suffer from at some point. Here we associate it with potential difficulties to formulate relevant questions.} We also provide the option to focus the question on a specific CDF report. This is particular important for questions that lack enough details and return valid answers from different reports. For example, asking a question about a vehicle or instrument that can be part of different missions can bring potential irrelevant results. Finally, SpaceQA also provides users with text snippets randomly extracted from CDF reports to stimulate question asking.

\begin{table}[htbp]
  \centering
  \caption{Example questions and the answers to them proposed by SpaceQA}
    \begin{tabular}{l}
        \toprule
    1. Which launcher will athena use? \\
     \hspace{3mm} Ariane 5 \\
    2. What wavelengths can be observed by NG-CryoIRTel? \\
    \hspace{3mm} 20-200µm \\
    3. What is the purpose of the tunable laser spectrometer?  \\
    \hspace{3mm} detect trace concentration of water and volatiles \\
    4. How is the ATHENA mirror structure manufactured?  \\
    \hspace{3mm} 3D-printing \\
    5. Where will NG-CryoIRTel be launched from?  \\
    \hspace{3mm} Tanegashima Space Centre \\
    6. Where is the panoramic camera mounted?  \\
    \hspace{3mm} a deployable mast \\
    7. When can dust storms occur during the MarsFAST mission? \\
    \hspace{3mm} at any time \\
    8. How long will the NG-CryoIRTel mission last?  \\
    \hspace{3mm} at least 5 years \\
    \bottomrule
    \end{tabular}%
  \label{tab:spaceqaexamples}
\end{table}

In table \ref{tab:spaceqaexamples} we show some example questions and the answers to those questions produced by SpaceQA. The system deals effectively with different types of wh-questions (what, which, where, when, why, and how), and provides appropriate answers in the form of instruments like rockets, units of measure, descriptions, locations, things, and time periods. Nevertheless, there are questions for which the system does not provide an answer, or the answer is wrong. While some of these are poorly specified questions\footnote{For example, the question \textit{Why can the sample material not be exposed to daylight?} does not add any information about what sample material they are referring to.} lacking details that can help to increase the confidence of the reader, others are just not properly answered. Recall that in our system we use models that were pre-trained  on general purpose corpora and fine-tuned on MS MARCO and SQUAD2.0, none of them rich in space. Therefore, a natural next step to improve the SpaceQA system is to generate a QA dataset for the space domain to fine-tuned the reader. 


\section{Generating quizzes to support training on quality management and assurance in space science and engineering}
\label{sec:spaceqquiz}

Quality management is a critical requirement to guarantee the success of space missions due to the complexity, cost, and risk, in occasions even for human lives, that they entail. Also to consistently produce space missions that meet the stakeholders expectations, ensuring that methods, processes, parts and materials are adequate and changes do not compromise results. ESA makes a continuous effort to train their staff in quality procedures and standards. Trainees are evaluated to determine the effectiveness of the training sessions, with quizzes as one of the main tools used in such evaluations. 

In this case study, we focus on SpaceQQuiz\footnote{SpaceQQuiz stands for {\it Space Quality Quiz}.}~\citep{garcia-etal2022}, a natural language generation system designed to help trainers to generate quizzes from documents describing quality procedures. Quality procedure documents cover topics like \textit{Anomaly and Problem Identification, Reporting and Resolution} or \textit{Configuration Management}, and include stakeholder responsibilities, activities, performance indicators and outputs, among others. 

Focused on text generation, like the previous one, this case study also falls mainly in the category of comprehension-intensive NLP projects described in section~\ref{sec:fw} and complements the open-domain question answering case study. It also leverages the same annotated dataset (SQuAD). However, in this case in addition to BERT-style transformer language models (used here to validate the quality of the generated questions) it relies on pre-trained generative language models for question generation. 

\begin{figure}[t]
    \centering
    \includegraphics[width=0.8\textwidth]{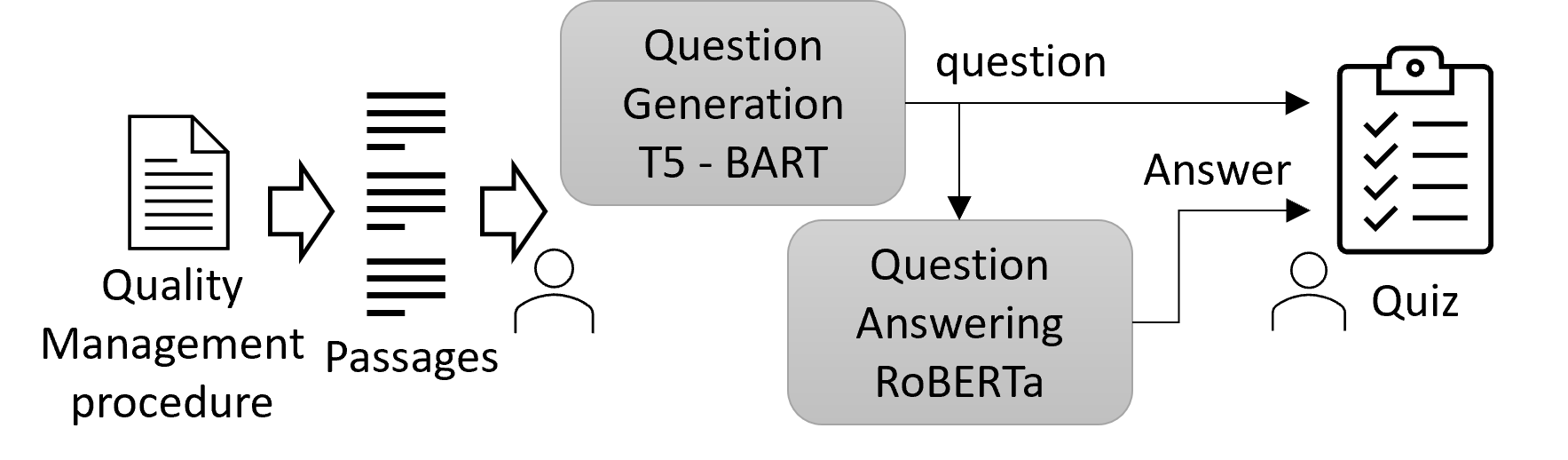}
    \caption{High-level architecture and main components of SpaceQQuiz.}
   \label{fig:qgarch}
\end{figure}

\subsection{The SpaceQQuiz system}
Figure \ref{fig:qgarch} shows the high-level architecture of SpaceQQuiz. A question generation model is run on each passage extracted from the document. The generated questions and the corresponding passages are fed to a question answering model that extracts the answer from the passage. Only questions with answers are included in the candidate list that is then refined by the trainer to generate the quiz.

The process starts when the trainer uploads a quality procedure document. SpaceQQuiz extracts the text from the PDF document using Apache PDFBox\footnote{Apache PDFBox \url{https://pdfbox.apache.org}} and uses regular expressions to identify  sections, subsections and paragraphs while removing non relevant text such as headers and footers or the table of content. The trainer is presented with a list of candidate sections so that she can choose the most interesting ones for the quiz. 

\subsubsection{Question generation}
We use state-of-the-art models based on transformers for question generation and question answering. Since we could not find specialized
models for the space or quality management domains, we reused models already pre-trained on general-purpose document corpora and fine-tuned on SQuAD.

To generate the questions we use a T5 model~\citep{raffelT52020}  and a BART model~\citep{lewis-etal-2020-bart} fine-tuned for question generation\footnote{Models reused from \url{https://github.com/patil-suraj/question_generation}}. We use two models in order to increase the number and variety of questions for each text passage. Both T5 and BART have excelled in sequence generation tasks, such as abstractive summarization and abstractive question answering. The models were fine-tuned using SQuAD1.1, 
which consists of 100K questions created from Wikipedia articles where answers are segments in text passages. 

T5 is fine-tuned using an answer-aware approach where the model is  presented with the answer and a passage to generate the question. T5 is trained on a multitask objective to i) extract answers, ii) generate questions for answers using passages as context, and iii) extract answers for the generated questions. Finally the answer for the generated question is compared with the answer used to generate the questions. BART is fine-tuned following an answer-agnostic approach where the model is trained to generate questions from passages without information about the answers. 

During generation, we use beam search as decoding method, with 5 as number of beams. Beam search keeps the most likely sequence of words at each time step and chooses the final sequence that has the overall highest probability. To avoid duplicity of questions in the final list, we compare them using cosine similarity between the question encoding generated through sentence transformers~\citep{reimers-gurevych-2019-sentence}. We discard questions similar to a previous one above an empirically defined threshold set at 0.8. 

\subsubsection{Question answering}
Once the questions have been generated we use a RoBERTa language model~\citep{liu2019roberta} fine-tuned for question answering on SQuAD2.0 
to extract answers from the passages. SQuAD2.0 adds 50,000 unanswerable questions to SQuAD1.1. Thus, the fine-tuned RoBERTa is able to generate answers or not depending on the question. If RoBERTa fails to generate an answer for a generated question we remove it from the candidate list of questions presented to the trainer.

\subsubsection{Quiz generation}
The trainer can select specific questions to include in the quiz by selecting them from the list of generated questions, answers, and passages displayed by the SpaceQQuiz user interface. Finally the system generates the quiz with a section containing only the questions to be handed to the trainee and another section reserved for the trainer with questions, answers and passages. 

\begin{table}[htbp]
\caption{Examples of questions generated by SpaceQQuiz.}
\centering
\resizebox{0.9\columnwidth}{!}{%
\begin{tabular}{p{\columnwidth}}
\toprule
What is the first source for raising a spacecraft Anomaly Report?               \\
\hspace{3mm} the spacecraft log is the first source for raising ... \\
What does the ARB have to do in case of  an anomaly detected in a shared infrastructure?                                                        \\
\hspace{3mm} notify the relevant infrastructure team                                         \\
Who can issue a supplier waiver?                                                \\
\hspace{3mm} OPS Project   Manager or Service Manager                                          \\
What does the leader of the operator's   team do with the raised Anomaly Reports? \\
\hspace{3mm} performs a preliminary review                                                     \\
Who chairs the   Software Review Board?                                           \\
\hspace{3mm} the owner of   the software,                                                      \\
What is   mandatory for the closure of a Problem Report?                          \\
\hspace{3mm} Root cause   identification                                                       \\
What are minor   non-conformances?                                                \\
\hspace{3mm} by definition, cannot be classified as major.                                   \\
\bottomrule
\end{tabular}%
}
\label{tab:qgquestions}
\end{table}

\subsection{Evaluation}
To evaluate SpaceQQuiz, we generate a quiz with 50 question-answer pairs from a quality procedure document titled \textit{OPS Procedure for Configuration Management}. Then, a quality management expert evaluates the generated questions using relevance and correctness as evaluation criteria. Table~\ref{tab:qgevaluation} reports the results of this manual evaluation.

In total 66\% generated questions are considered relevant and grammatically correct and 60\% of the answers are also regarded as correct by the evaluator. If we focus only on the answers of relevant and correct questions then the percentage of accurate answers improves to 81.8\%. The level of accuracy for the question generation still requires to keep an human in the loop in order to guarantee the quality of the questions in the quiz. Ultimately, it is the responsibility of the domain expert to decide upon the selection of questions to be included in the quiz. 

By analysing incorrect question and answer pairs, we realize that despite being grammatically correct and relevant, some questions are just not possible to answer from the context used to generate them (see for example questions 1 and 2 in table~\ref{tab:qgwrongquestions}). This is consequence of a failure in the question answering module that produces an answer for such questions. Another example of wrong functioning of the question answering module is shown in question 3 in table~\ref{tab:qgwrongquestions}, where the answer to the given question is extracted from the example in round brackets. A possible solution for this case could be to discard  examples in the text before feeding the question generation and the question answering modules. Note that the question answering model we are using is only able to extract possible answers from text, not to generate new text that may answer the question. 

A generative question answering model informed by the question generation model by jointly training both models could be a better way to address this type of errors. Similar approaches, like~\citep{wan-bansal-2022-factpegasus}, have been followed to prevent hallucination in abstractive text summarization tasks, for example. 

\begin{table}[htbp]
  \centering
    \caption{Evaluation of the question generation and question answering modules by a quality assurance expert. * Indicates that only answers with a valid question are evaluated.}
    \begin{tabular}{lr}
    \toprule
          & \multicolumn{1}{l}{Accuracy} \\
    \midrule
    Generated questions & 0.660 \\
    Extracted Answers & 0.600 \\
    Extracted Answers* & 0.818 \\
    \bottomrule
    \end{tabular}%
  \label{tab:qgevaluation}%
\end{table}%

\begin{table*}[t]
  \centering
  \caption{Example questions evaluated as incorrect. In bold, the answers extracted by the question answering module.}
  \resizebox{0.9\textwidth}{!}{%
    \begin{tabular}{cp{0.1\textwidth}p{0.8\textwidth}}
    \toprule
    \multirow{2}{*}{1} & Context & In the process of \textbf{configuration identification} the team shall be aware on what is needed to be put under configuration control. \\
    & Question & What shall the team know on what is needed to be put under configuration control? \\
    \midrule
    \multirow{2}{*}{2} & Context & In a \textbf{continuous service} there is the concept of living baseline over a dynamic scope. \\
    & Question & What is the concept of living baseline over a dynamic scope? \\
    \midrule
    \multirow{2}{*}{3} & Context & Item configuration, in terms of implemented functions (e.g. \textbf{software version 2.0)} \\
    & Question & What is item configuration in terms of implemented functions? \\
    \midrule
    \multirow{2}{*}{4} & Context & The system under configuration includes also \textbf{the items received as Customer Furnished Item.} \\
    & Question & What does the system under configuration include? \\
    \bottomrule
    \end{tabular}%
  }
  \label{tab:qgwrongquestions}%
\end{table*}%

For some correct and relevant questions, see, e.g., question 4 in table \ref{tab:qgwrongquestions}, the question answering module just returns partial answers. In this case the word \textit{also} means that the answer in this context complements the answer already provided in another text excerpt. This a limitation of the extractive question answering module since it only extracts consecutive sequence of tokens from text passages as answers.

Finally the domain expert evaluator reported that in some cases the problem is related to the source text used to generate the question, which may not be clear enough to formulate appropriate questions. Thus, wrong questions might indicate text excerpts that need to be reviewed by the authors of the quality management document to convey their message more clearly.
 
\section{Information extraction for Long-Term Data Preservation in space}
\label{sec:fair}


The large amount of new space missions in areas like Earth Observation planned for the next years will lead to a major increase of space data that adds up to the data legacy of current and past missions. Together with the growing demands from the user community, this marks a challenge for satellite operators, space agencies and data providers regarding the coherent preservation and optimum availability and accessibility of the different data products. Among the main goals of the European EO Long Term Data Preservation Framework,\footnote{LTDP Introduction and Objectives (\url{https://earth.esa.int/eogateway/activities/gscb-and-ltdp/ltdp-introduction-and-objectives})} the need to ensure and facilitate data accessibility and usability is a key one. To achieve this goal it is necessary to enhance the ability of machines to automatically find and use scientific information from related disciplines in addition to supporting reuse by individuals. Text analytics systems can contribute to such vision by automatically extracting information from relevant sources like scholarly communications, technical reports, mission feasibility studies, design documents or mission reports, exposing such information as machine-readable metadata that facilitates discovery through automated means. Once available, such metadata is instrumental for the development of information retrieval systems, such as search and recommendation engines, facilitating access.

As shown in table~\ref{tab:ucdim}, this case study focuses on information extraction and follows a knowledge-based approach to text analytics. We present the methodology followed to extract domain-specific terminology and its integration in a pre-existing, general-purpose knowledge graph, extending and customizing it for Earth and Environmental sciences,\footnote{Both relevant for Earth Observation, the main strategic area of interest for this study.} and supporting the development of domain-specific text analytics services for information extraction. We also illustrate how the metadata extracted by such services can be leveraged by powerful search and recommendation engines. 
We use expert.ai technology\footnote{Expert.ai core technology: \url{https://www.expert.ai/products/technology}} for natural language processing and understanding, which relies on a general-purpose lexico-semantic knowledge graph with approximately 400K lemmas, 300K concepts, and 80 different types of relations, rendering 3 million links between concepts. 

\subsection{Text resources for terminology extraction and model training}
A domain-specific text corpus is necessary to adapt  existing text mining tools to the vocabulary used by the target users
. First, we worked with documental sources facilitated by ESA, including proceedings of ESA-sponsored conferences like Big Data from Space\footnote{BIDS (\url{https://www.bigdatafromspace2021.org/})} and the PV conference series\footnote{\url{https://earth.esa.int/eogateway/activities/gscb-and-ltdp/pv-conferences}} about ensuring long-term preservation and adding value to scientific and technical data,  as well as the ESA corporate taxonomy, the glossary of long-term preservation of earth
observation space data\footnote{\url{https://ceos.org/document_management/Working_Groups/WGISS/Interest_Groups/Data_Stewardship/White_Papers/EO-DataStewardshipGlossary.pdf}}, and the ESA Technology Tree.\footnote{\url{https://www.esa.int/About_Us/ESA_Publications/STM-277_ESA_Technology_Tree}} We extract and process the text from such documents using expert.ai and the pre-existing version of its knowledge graph. By comparing the information thus extracted, e.g. keywords, multi-word expressions, concepts, lemmas, entities, with the terms captured by the expert.ai knowledge graph we were able to identify a vocabulary gap (see section~\ref{terminologyAnalysis} for a detailed description of the terminology analysis). This resulted in the extension of the knowledge graph with 579 new acronyms and their definitions, 77 departments in the ESA corporate organization, over 100 new lemmas and multi-word expressions, 83 space missions, 16 organizations, over 200 proper nouns (prominent public figures, such as authors and scientists), and 186 other terms, as well as the explicit relations between them and the corresponding concepts in the knowledge graph. 

In a second stage we continued extending the knowledge graph with publicly available information from additional scientific publications and set to collect a public documental corpus. We focused on Springer Nature's SciGraph,\footnote{\url{https://www.springernature.com/gp/researchers/scigraph}}. Other sources considered include open access publications in Earth and Environmental sciences from OpenAire\footnote{\url{https://www.openaire.eu}} and Scopus.\footnote{\url{https://www.scopus.com/}}. SciGraph is a knowledge graph of scholarly communications covering funding agencies, research projects, conferences, affiliations, and publications. The main source of information for SciGraph is the Springer Nature editorial group, which ensures high quality data from trusted and reliable sources. Publications include journals, articles, books, and book chapters from the last 200 years. 


SciGraph uses the {\tt schema:about}\footnote{\url{http://schema.org/about}} property to relate publications to the Fields of Research classification\footnote{\url{https://www.arc.gov.au/grants/grant-application/classification-codes-rfcd-seo-and-anzsic-codes }} (FOR). FOR includes major fields and related sub-fields of research and emerging areas of study investigated by businesses, universities, national research institutions and other organizations. FOR is a taxonomy with three levels: divisions (2 digits), groups (4 digits) and fields (6 digits). Each division is based on a broad discipline. Groups within each division are those which share the same broad methodology, techniques and/or perspective as others in the division. Each group is a collection of related fields of research. Groups and fields of research are categorized to the divisions sharing the same methodology rather than the division they support. An example of division, group and field of research hierarchy is: 09 Engineering (division), 0901 Aerospace Engineering (group), 090101 Aerodynamics (field). The FOR taxonomy totals 22 divisions. 

Focusing on the domain-specific vocabulary relevant for Earth and Environmental sciences, we conducted a survey within a group of 12 vulcanologists, sea observation scientists and climatologists to determine: i) the most relevant fields of research for their work and ii) the journals and venues where them and their peers publish their scientific contributions. We generated a spreadsheet with the FOR taxonomy and asked our scientists to mark the most relevant fields for their work, following these guidelines: First, select the most relevant division associated with your research; then determine the most relevant group within that division; and finally identify the most relevant field within that group. We also advised them to prioritize research fields that overlap most with their research, e.g. Vulcanology, over research fields related to specific techniques used in their work, e.g. Analytical Chemistry. 

The groups and fields of research corresponding to the Earth Sciences division selected by our team of experts was the following:
\begin{itemize}
    \item \textbf{Atmospheric sciences}: Atmospheric Aerosols, Atmospheric Dynamics, Atmospheric Radiation, Climate Change Processes, Climatology (excl. Climate Change Processes), Cloud Physics, Meteorology, Tropospheric and Stratospheric Physics.
    \item \textbf{Geochemistry}: Exploration Geochemistry, Inorganic Geochemistry, Isotope Geochemistry, Organic Geochemistry.
    \item \textbf{Geology}: Basin Analysis, Extraterrestrial Geology, Geochronology, Igneous and Metamorphic Petrology, Marine Geoscience, Mineralogy and Crystallography, Ore Deposit Petrology, Paleontology (incl. Palynology), Petroleum and Coal Geology, Sedimentology, Stratigraphy (incl. Biostratigraphy and Sequence Stratigraphy), Structural Geology, Tectonics, Volcanology.
    \item \textbf{Geophysics:} Electrical and Electromagnetic Methods in Geophysics, Geodynamics, Geophysical Fluid Dynamics, Geothermics and Radiometrics, Gravimetrics, Magnetism and Paleomagnetism, Seismology and Seismic Exploration.
    \item \textbf{Oceanography:} Biological Oceanography, Chemical Oceanography, Physical Oceanography.
    \item \textbf{Physical Geography and Environmental Geoscience:} Geomorphology and Regolith and Landscape Evolution, Glaciology, Hydrogeology, Natural Hazards, Palaeoclimatology, Quaternary Environments, Surface Processes, Surfacewater Hydrology.
\end{itemize}

Similarly, for Environmental Sciences:
\begin{itemize}
    \item \textbf{Ecological applications:} Ecological Impacts of Climate Change, Ecosystem Function, Invasive Species Ecology, Landscape Ecology.
    \item \textbf{Environmental Science and Management:} Aboriginal and Torres Strait Islander Environmental Knowledge, Conservation and Biodiversity, Environmental Education and Extension, Environmental Impact Assessment, Environmental Management, Environmental Monitoring, Environmental Rehabilitation (excl. Bioremediation), Maori Environmental Knowledge, Natural Resource Management, Pacific Peoples Environmental Knowledge, Wildlife and Habitat Management.
    \item \textbf{Soil sciences:} Carbon Sequestration Science, Land Capability and Soil Degradation, Soil Biology, Soil Chemistry (excl. Carbon Sequestration Science), Soil Physics.
\end{itemize}

\subsection{Corpus generation}
We generate our corpus by applying two filters to the SciGraph Articles Dump: i) Publication Date and ii) Fields of Research. From the SciGraph articles dump, a compressed file with 10,653 JSON files with all the articles, we focus on a subset containing those  published since 2016 that belong to the Earth Sciences or Environmental Sciences fields of research, and produce a single JSON file with the title and  abstract of such articles. The JSON file contains 49.693 articles, with 13M tokens, among which 271K are unique. 61\% of them (30.190 articles) are labeled as Earth Sciences papers, while the rest (19.503 articles) belong to the Environmental Sciences field. The main subcategories in Earth sciences (36\%) are Geology and Physical Geography (25\%), while the remaining subcategories are uniformly distributed over the rest of the sample. For Environmental sciences, Environmental Science Management (55\%) and Soil sciences (43\%) are dominant.\footnote{The corpus, along with the  metadata extracted from it, is available in Zenodo \url{https://zenodo.org/record/4721343/files/scigraph_corpus_zenodo.json}.}

\subsection{Terminology analysis}\label{terminologyAnalysis}
The analysis of the corpus is useful to assess and extend terminology coverage in the domains of interest. To analyze the corpus we use a pre-existing version of the text mining and enrichment services based on expert.ai technology to detect concepts that are not already encoded in our knowledge graph. Such concepts, along with multi-word expression also detected by the text analytics engine, are used to enrich the knowledge graph. In addition, named entities like people, organization and places are manually inspected to detect errors and improve the accuracy of the named-entity recognition (NER) module. Finally, we carry out a weirdness index analysis to detect words that are specific of the target scientific domains. 

The expert.ai knowledge graph is a semantic network that represents knowledge as a graph of concepts and relationships between them. The nodes of this knowledge graph are called syncons, and they are linked to each other through semantic and linguistic relationships in a hierarchical structure. Each syncon has a main lemma, which is a canonical representations of words and collocations without conjugation, number or gender. The complete meaning of a word or expression in the text comes as a combination of its main elements after disambiguation (grammar type, syncon, definition/gloss, domain, and frequency relations), as well as the different types of connections, e.g. hypernymy, hyponymy, it may have with other syncons. This explicit representation results in a greater ability to understand language, which can be adapted to a specific domain by adding new concepts and relations that enrich the metadata extraction and improve text comprehension. In our case, we extend a general-purpose version of the graph with new terms from the corpus, selecting those that are more representative of the vocabulary used by our target communities 

One of the advantages of this approach is that the knowledge graph can be used to disambiguate the meaning of a word by recognizing its context. This disambiguation process comprises several phases of analysis including a lexico-grammatical analysis, which identifies e.g. nouns, proper nouns, and verbs, a syntactical analysis that identifies word groups at different levels e.g. noun phrases and verb phrases, and a semantic analysis, which finally determines the meaning of each document token according to the knowledge graph. Additionally, named-entity recognition allows spotting names referenced within the text, such as proper nouns, organizations, and locations. 

We process the corpus and extract metadata from the documents by feeding the text mining services with the title and abstract of each paper. We generate the following metadata: 
\begin{itemize}
    \item \textbf{Domain:} Field(s) of knowledge, based on main concepts.
    \item \textbf{Organizations:} Organization names or aliases.
    \item \textbf{People:} Person names or aliases.
    \item \textbf{Places:} Places names or aliases.
    \item \textbf{Known Concepts:} Concepts found in the text and present in the knowledge graph.
    \item \textbf{Concepts:} Concepts in the document that are not in the graph.
    \item \textbf{Main Syncons:} Most relevant concepts mentioned in the text that are represented in the graph. 
    \item \textbf{Main Groups:} Most relevant noun phrases and multi-word expressions in the text. 
    \item \textbf{Main Lemmas:} Most frequent lemmas found in the text.
    \item \textbf{Main Sentences:} Most relevant sentences found in the text.
\end{itemize}

To identify the corpus terminology that was not covered yet by the knowledge graph, we focused on the lemmas of words and multi-word expressions (noun phrases) that the text mining service was not able to associate with a concept in the knowledge graph. Table~\ref{tab:fairlemmas} shows the top 10 most frequent of such lemmas. As shown in the table, most of the unknown terms are chemical compounds and measures, which is not surprising since the knowledge grtaph was not originally conceived to cover Chemistry.

\begin{table}[ht!]
    \resizebox{0.5\textwidth}{!}{
    \begin{minipage}{.5\linewidth}
    \centering
     \caption{Top 10 lemmas  without a concept in the knowledge graph}
      \label{tab:fairlemmas}
      \begin{tabular}{cc}
        \toprule
        Lemma & Count \\
        \midrule
        ha-1 & 543 \\
        kg-1 & 444 \\
        R2  & 426 \\
        NO3 & 414 \\
        18O & 387 \\
        sea surface temperature & 373 \\
        Mw & 343 \\
        SiO2 & 318 \\
        CMIP5 & 305 \\
        m-2 & 263 \\
      \bottomrule
    \end{tabular}
    \end{minipage}%
    }
    \resizebox{0.5\textwidth}{!}{
    \begin{minipage}{.5\linewidth}
    \centering
      \caption{Top 10 multi-word expressions in the Corpus}
      \label{tab:fairgroups}
      \begin{tabular}{cc}
        \toprule
        Main group & Count \\
        \midrule
        soil sample & 641 \\
        species richness & 316 \\
        soil moisture & 296 \\
        climate model & 272 \\
        groundwater sample & 266 \\
        soil property & 264 \\
        land use & 263 \\
        ecosystem service & 233 \\
        climate variability & 214 \\
        soil fertility & 200 \\
      \bottomrule
    \end{tabular}
    \end{minipage}%
    }
\end{table}


Another source of potential concepts to integrate in the knowledge graph are the main Groups or multi-word expressions detected by the software. Main groups in table~\ref{tab:fairgroups}, contain phrases of nouns, verbs and prepositions. Similar to lemmas, after identifying candidate expressions to be included in the graph, a knowledge engineer needs to determine whether a phrase can be represented as a concept and the exact location of the graph and form in of such representation. This will depend on different factors, including the possibility of explicitly linking the new concept with existing concepts through relations or the existence of previous related concepts in the hierarchy.


Table~\ref{tab:fairner} shows the top 10 most frequent named entities found in the corpus. Since these are the most frequent entities, there is little chance of error. However, proper nouns like {\it Forest} could be further investigated to see if it actually refers to a person or not. Among the organizations, {\it sea surface temperature} is clearly an error that needs to be addressed. The reason for such error is that {\it sea surface temperature} is not explicitly encoded as part of the knowledge graph and hence the entity type needs to be inferred. In this type of situations, the knowledge engineer needs to decide, based on the relevance of the entity, whether to include it explicitly in the graph.

\begin{table}[ht!]
  \centering
  \caption{Top 10 named entities in the corpus per frequency}
  \label{tab:fairner}
\resizebox{\textwidth}{!}{  
  \begin{tabular}{p{1.25cm}c|p{2.5cm}c|p{4.5cm}c}
    \toprule
    Person & Count & Place & Count & Organization & Count\\
    \midrule
    Salvatore Pinto &	315	& China	& 4,292	& European Community &	280 \\
    Shannon & 135	& India	& 2,074	& European Union	& 247 \\
    Ma	& 53 &	United States of America & 1,366 &	Intergovernmental Panel on Climate Change	& 152 \\
    Biochar & 	50	 & Iran	& 972	& Cd	& 144 \\
    Linnaeus	& 50	& Europe	& 838	& International Union for Conservation of Nature &	136 \\
    Anne &	48 &	Japan &	757 &	soil organic carbon &	129 \\
    April &	46 &	Brazil &	710 &	ECMWF &	115 \\
    Pb & 	38	& Atlantic Ocean	& 687	& World Health Organization	& 106 \\
    Forest &	35 &	Mediterranean Sea &	624 &	O2 plc &	98 \\
    Rossby &	30 &	Italy &	614 &	sea surface temperature &	82 \\
  \bottomrule
  \end{tabular}
 }
\end{table}

Table~\ref{tab:fairstats} shows the number of words per metadata type shown by our analysis to be previously known or unknown in the knowledge graph Note that not all terms need to be included. For example, only named entities classified under the wrong entity type need to be integrated in the graph so that the disambiguation process has more information about them when determining the correct entity type. The rest of unknown entities correctly classified does not need to be added. Similarly, only unknown lemmas and groups that are ambiguous need to be integrated in the graph so that they can be disambiguated properly. Since there is a considerable number of unknown lemmas, groups and entities, we apply the Pareto principle and focus on the 20\% most frequent words for each metadata type. This subset of words is handed to a team of knowledge engineers and linguists in charge of their integration in the expert.ai knowledge graph. In total, the knowledge engineers need to analyze and process 5,070 words, with an estimated total effort of 2.5 person months. 

\begin{table}[ht!]
  \centering
  \caption{Number of known vs. unknown corpus terms }
  \label{tab:fairstats}
  \resizebox{\textwidth}{!}{  
  \begin{tabular}{ccccc}
    \toprule
     & Known & Unknown & 20\% Most frequent unknown & Total\\
    \midrule
    Lemmas	& 22,978 &	26,974	& 556	& 49,952 \\
    Groups &	735 &	171,047	& 3,121	& 171,782 \\
    Persons &	1,463	& 6,520	& 341 & 7,983 \\
    Places & 6,371	& 20,853	& 542 &	27,224 \\
    Organizations &	1,730 &	21,755 &	510 &	23,485 \\
  \bottomrule
\end{tabular}
}
\end{table}

One way to streamline the integration of new terminology in the knowledge graph is to prioritize those terms that are more specific of the reference corpus over those that are more generalistic. To this purpose, similarly to~\citep{Berquand2020SpaceMD}, we apply Weirdness Index~\citep{10.1007/11575801_25} filtering to rank the candidate terms. The Weirdness Index allows comparing the use of a word, based on its frequency, between a domain-specific corpus and a large corpus representing general-purpose language. In this case, we use the British National Corpus\footnote{\url{http://www.natcorp.ox.ac.uk}} (BNC) as our general corpus. We calculate the Weirdness Index as shown in equation~\ref{eqn:wi}, where $f_S$ is the frequency of the word in the specialized corpus, $f_G$ its frequency in the general corpus, and $N_S$ and $N_G$ are the number of tokens in the specialized and in the general corpus, respectively. Table~\ref{tab:fairwi} shows some of the terms with the highest and lowest Weirdness Index in our corpus.

\begin{equation}
\label{eqn:wi}
    W = \frac{N_{G}f_S}{(1+f_G)N_S}
\end{equation}

\begin{table}[ht!]
  \centering
  \caption{Top 10 terms with the highest and lowest weirdness index in the corpus}
  \label{tab:fairwi}
  \begin{tabular}{ccccc}
    \toprule
    Highest WI & Weirdness index & Lowest WI & Weirdness index \\
    \midrule
    ENSO & 17,250.2 & Win & 0.0043 \\
    N2O & 9,480.3 & Money & 0.0042 \\
    CMIP5 &  7,443.28 & Studio & 0.0039 \\
    Modelling & 5,665.83 & Pupil & 0.0039 \\
    WRF & 4,847.59 &  Terry & 0.0037 \\
    NDVI &  4,598.33 & Worry & 0.0033 \\
  \bottomrule
\end{tabular}
\end{table}

\subsection{Exploiting the extracted metadata}
\begin{figure}[t]
    \centering    \includegraphics[width=\textwidth]{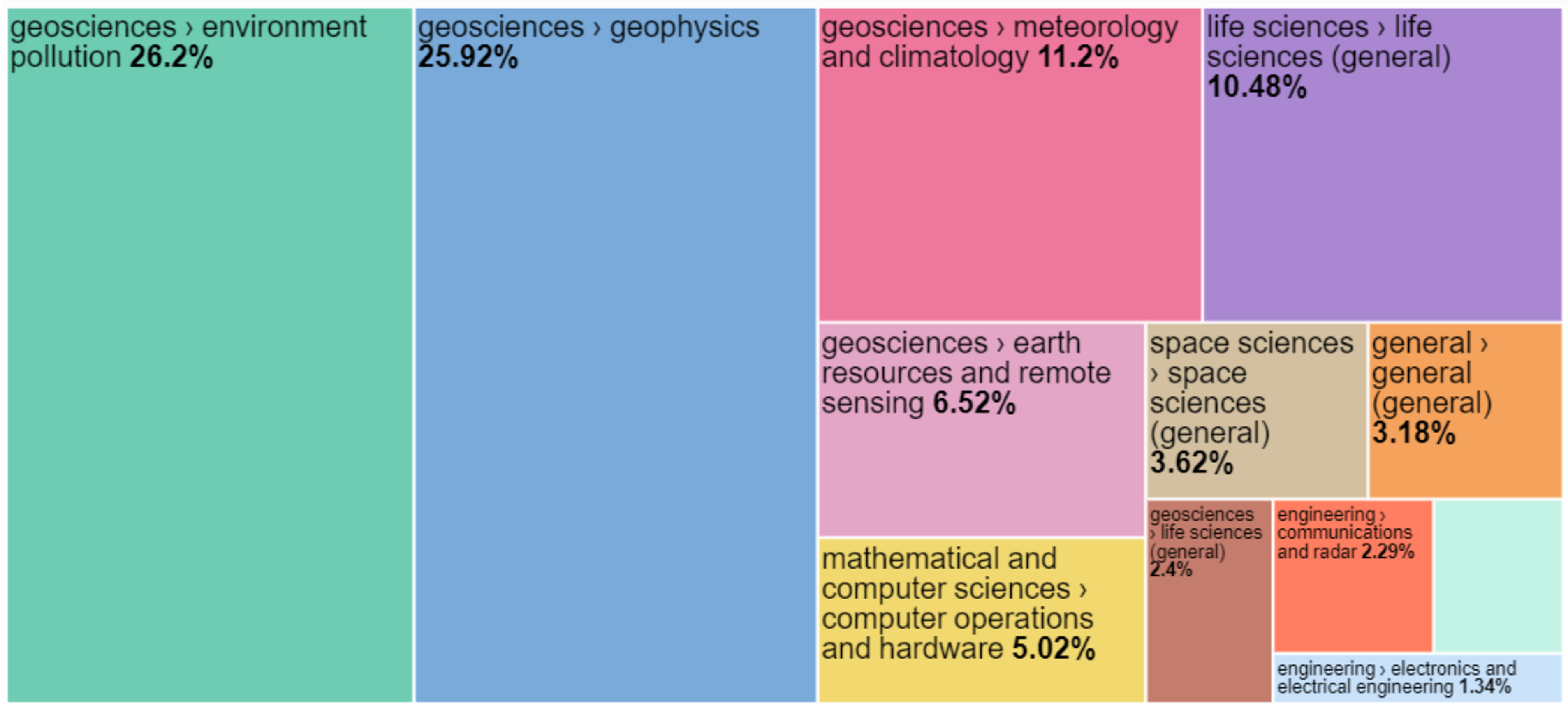}
    \caption{Fragment of dashboard with the distribution of metadata over the corpus.}
   \label{fig:fairdash}
\end{figure}

Once the knowledge graph has been extended and adapted to the specific domain, any collection of documents can be processed to extract metadata\footnote{A live demo illustrating the semantic metadata presented in this paper that can be extracted from Earth and Environmental sciences documents can be found at: \url{https://reliance.expertcustomers.ai/enrichment}} from them, which can then be used to improve access to such information. This type of application is illustrated in figure~\ref{fig:fairdash}, which shows part of an example dashboard\footnote{The interactive dashboard is available at: \url{https://reliance.expertcustomers.ai/dashboard/app/r/s/HRk3f} user/pass: guest/relish2022!} built with Kibana\footnote{\url{https://www.elastic.co/kibana}} 
to visualize and explore a document collection based on the distribution of the information extracted from it as semantic metadata. 

The services described in this section are also available at the European Open Science Cloud (EOSC)\footnote{\url{https://marketplace.eosc-portal.eu/services/enrichment-api/information}} and are currently used among others by ROHub,\footnote{\url{https://reliance.rohub.org}} an online platform that aims at managing, preserving, and providing access to research work, including scientific data, code, and literature, in order to extract information from research objects in a variety of scientific communities, which currently include among others Astrophysics and Bioinformatics, as well as Earth and Environmental sciences.

The resulting metadata can also be used to enhance search and recommendation engines, alleviating some of the limitations of keyword-based approaches, including query ambiguity and lack of semantics. Keyword-based search engines may miss documents that contain synonyms of query keywords and morphological variations such as verb conjugations or even plurals, with an impact on recall. By leveraging semantic metadata generated as proposed above, where each concept identifies uniquely a word as well as other semantically related terms like synonyms and hyponyms, search and recommendation engines can be better equipped to deal effectively with ambiguity. Examples of this type of systems include the Collaboration Spheres,\footnote{\url{https://reliance.expertcustomers.ai/spheres}} a search-by-example system whose evaluation~\citep{Rico2017CollaborationSA} showed the benefits of this approach to explore large collection of scientific documents, reducing the cognitive load associated to this task.


\section{Assisted evaluation of the innovation potential of OSIP ideas}
\label{sec:osip}
In the previous section we showed how to apply the NLP framework presented in section~\ref{sec:fw} to extract information in the form of semantic metadata from documents in scientific areas that are relevant for space operations, like Earth and Environmental Sciences, contributing to long-term data preservation in space. In this section, we focus on a particular application of such metadata to produce, based on the semantic similarity of new ideas submitted to the OSIP\footnote{The Open Space Innovation platform (\url{https://ideas.esa.int})} platform with previously funded ideas, studies or projects, a score that quantifies how innovative such idea can be. 

As shown in table~\ref{tab:ucdim}, this case study also focuses on information extraction, However, the main goal is to perform comprehension tasks related to understanding and comparing ideas with other ideas and previous work. Since the innovation score needs to be justifiable, the explainability aspect is also important. We did not have access to annotated datasets for model training, which in addition to the previous factors advised for a knowledge-based approach to address the language understanding challenges in this case study.

This initiative had a twofold objective. First, to support the evaluation of ideas received by the OSIP platform through an AI system based on NLP, enabling internal collaboration within ESA and with ESA Member State delegates, the space industry, universities, and other organizations. And second, to deploy the service in production, minimizing operational costs by reusing pre-existing infrastructure and services. To achieve such objectives, the following capabilities were deployed at ESA: i) the ingestion of documents and their enrichment with metadata extracted from their content. The type of documents supported included ideas and campaigns hosted at the HYPE database of the OSIP platform, General Studies Program\footnote{\url{https://www.esa.int/Enabling_Support/Preparing_for_the_Future/Discovery_and_Preparation/About_the_GSP}} (GSP) studies from the GSP database, and projects funded by the European Commission under FP7 and Horizon 2020 programmes; ii) the calculation of a novelty score between 0 and 100 to allow for every idea under evaluation to be compared against other ideas previously selected, GSP study descriptions, and projects funded by the European Commission; and iii) the extension of the description of each idea in HYPE with its novelty score and the metadata used to calculate it, providing evidence-based means to explain such score.

\begin{figure}[ht]
    \centering
    \includegraphics[width=1.0\textwidth]{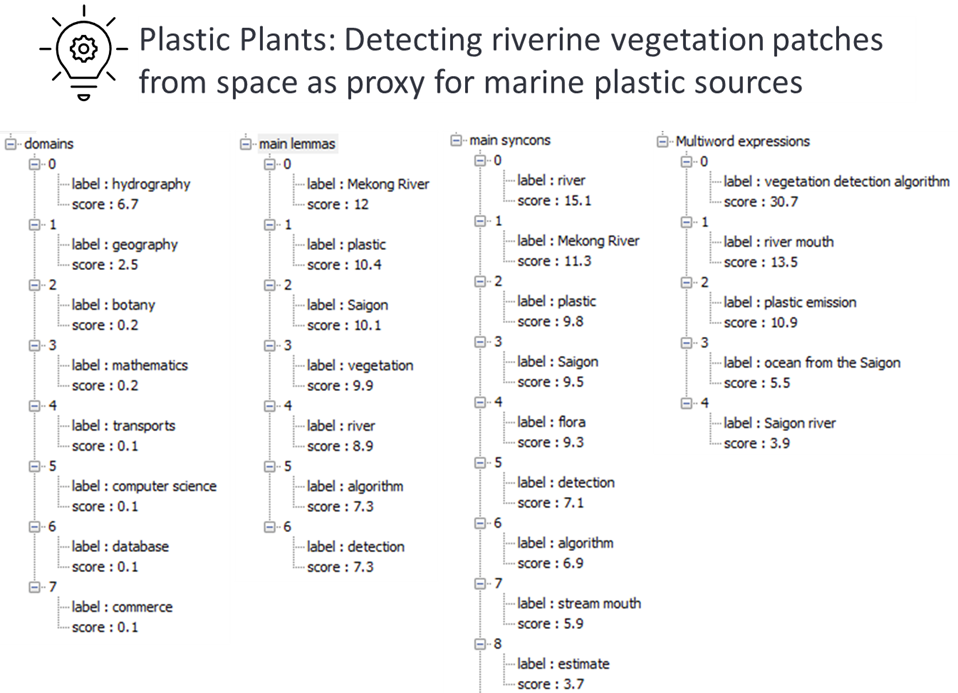}
    \caption{Example of metadata extracted from OSIP ideas.}
   \label{fig:osipmd}
\end{figure}

To improve the generated metadata, the coverage of space-related terminology in the expert.ai knowledge graph was further extended, continuing the work described in section~\ref{sec:fair}. 
Figure~\ref{fig:osipmd} illustrates the metadata extracted for one of the ideas submitted to the OSIP platform.

\subsection{Architecture and workflow}
\begin{figure}[t]
    \centering
    \includegraphics[width=1.0\textwidth]{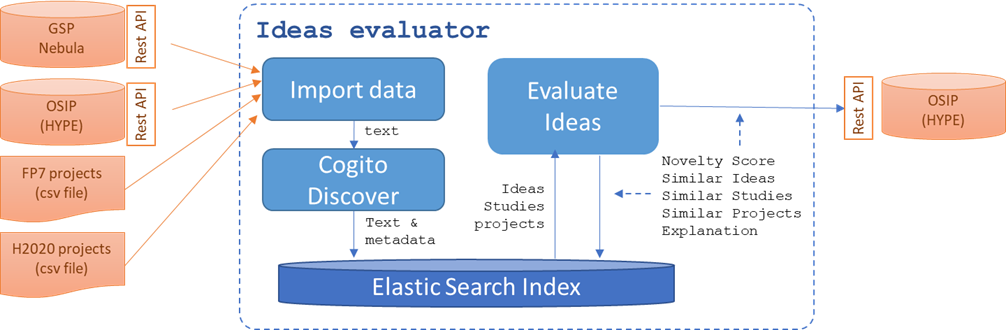}
    \caption{OSIP novelty evaluation service architecture.}
   \label{fig:osiparch}
\end{figure}

The architecture of the OSIP novelty evaluation service is depicted in figure~\ref{fig:osiparch}. The import module ingests data from the different sources considered (the OSIP platform, to extract ideas and campaigns; the Nebula\footnote{\url{https://nebula.esa.int}} library containing studies; and documentation about the projects funded by the FP7 and H2020 EC programs\footnote{Available from: \url{https://cordis.europa.eu/projects/en}}). The text that is extracted from such sources, corresponding to ideas, campaigns, studies, and projects, is sent to an instance of the expert.ai text analytics services hosted at ESA's facilities (Cogito Discover) that extracts semantic metadata from it using the extended knowledge graph. The resulting metadata is then stored in an Elasticsearch index along with the original text of the document plus the metadata that was extracted from each data source. The fields indexed in Elastichsearch include among others {\tt  Id\_}, {\tt Title}, {\tt Description}, {\tt StartDate}, {\tt EndDate}, {\tt Keywords} and {\tt AssociatedDocuments}, as well as the metadata extracted by Cogito Discover: {\tt Domains}, {\tt Organizations}, {\tt People}, {\tt Places}, {\tt Concepts}, {\tt MainGroups}, {\tt MainLemmas}, {\tt MainSentences}, and {\tt MainSyncons}. Next, the novelty of each idea is evaluated based on the similarity of the idea under evaluation with other ideas that were previously selected, implemented or archived and executed studies, as well as previous FP7 and H2020 funded projects. The novelty score plus the documents similar to the idea are saved in the Elasticsearch index. Finally, once annotated, the ideas are pushed back into OSIP, including the novelty evaluation score, its associated metadata, and the related documents.

\subsection{Novelty score evaluation}
To assess the novelty of an idea we propose to calculate the similarity between the idea and other ideas previously selected, implemented or archived, as well as studies and EC projects, so that the evaluator can count with this information as a first step towards the final decision. To this purpose, we use a simple metric based on the intuition that if an idea is very similar to other idea, study or project the novelty score should be low. Also the other way around, ideas that are different from previous activities are considered to be more novel. 

Unlike metrics for text similarity based exclusively on either lexical (keyword-based) or semantic similarity~\citep{10.5555/1597538.1597662}, we leverage all the information that has been previously indexed in Elasticsearch, which in our case includes both textual fields and semantic metadata extracted by the Cogito Discover module. This allows us to leverage the native similarity evaluation capabilities of Elasticsearch, which work directly with the values of the indexed fields. More specifically, we focus the calculation of the similarity between an idea and other documents on the information indexed about main lemmas and main concepts.  Other key terms are obtained from titles and descriptions by selecting the words with the highest TF-IDF scores. Equation~\ref{eqn:nov} defines the novelty score of an idea \textit{i}, based on its Elasticsearch similarity (sim) with  a collection of other ideas \textit{I}, a collection of studies \textit{S}, and a collection of projects \textit{P}.

\begin{equation}
\small
\label{eqn:nov}
noveltyScore(i,I,S,P) = 100*(1-max\{sim(i,I), sim(i,S), sim(i,P)\})
\end{equation}

The novelty score evaluation generates the following metadata for the ideas under evaluation: 
\begin{itemize}
    \item \textbf{noveltyCalculated:} Whether the novelty score was computed.
    \item \textbf{noveltyScore,} from 0 to 100. The higher, the more novel the idea. 
    \item \textbf{similarIdeas:} Similar ideas found for the evaluated idea.
    \item \textbf{similarProjects:} Similar studies and FP7 and H2020 projects.
\end{itemize}

\begin{figure}[ht!]
    \centering
    \includegraphics[width=0.87\textwidth]{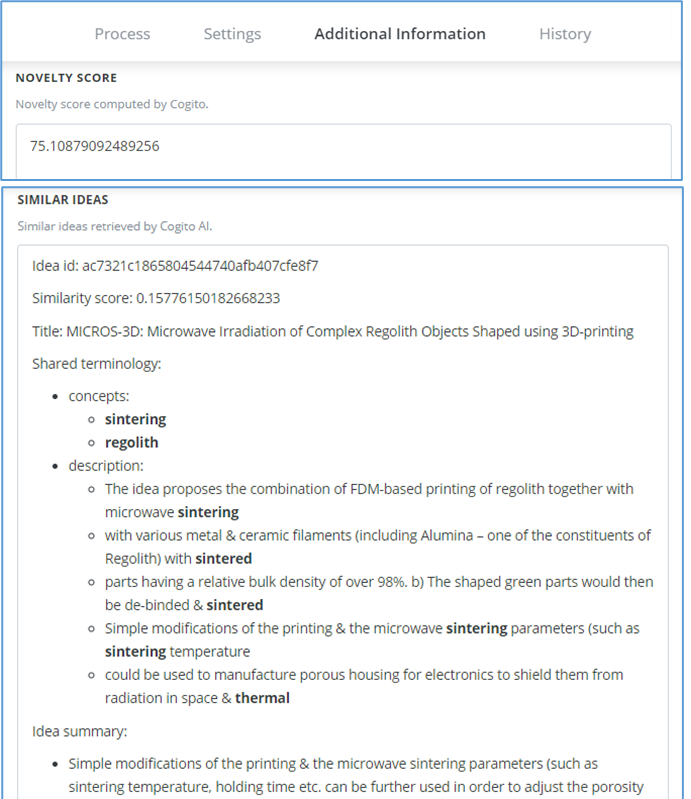}
    \caption{Screenshot of Novelty Score evaluation in OSIP.}
   \label{fig:osipscore}
\end{figure}

The {\tt similarIdeas} and {\tt similarProjects} metadata contains  the list of similar documents and provides information that \textbf{explains} the similarity with the evaluated idea by explicitly linking them. 
On a daily basis, every idea under evaluation is updated with novelty score and its related metadata. Such metadata can be viewed in the OSIP platform by selecting an idea, navigating to the menu option “Manage Idea”, and clicking on “Additional Information”. Figure~\ref{fig:osipscore} shows a screenshot taken from the production environment of the OSIP platform for an idea (with novelty score 75.1), for which the system has identified a slightly similar idea\footnote{\textit{Demonstration of radiation / thermal shielding with (small scale) inflatable gas tank + regolith sintered (with solar lens) structure.} (\url{https://tinyurl.com/mt77ypd8})} previously funded by OSIP and highlights the semantic metadata both ideas have in common, as well as the similarity score between them.


\begin{figure}[ht!]
    \centering
    \includegraphics[width=\textwidth]{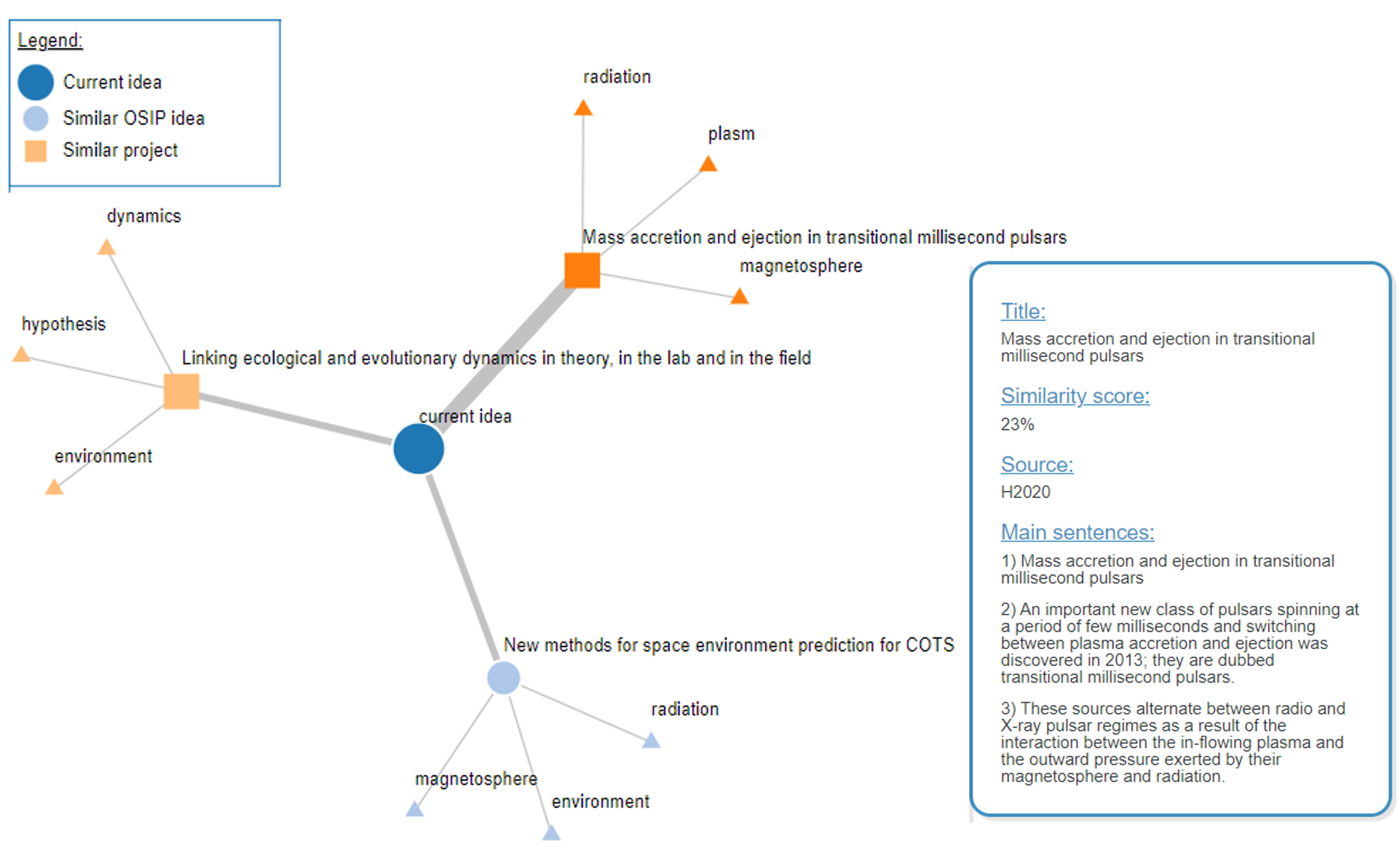}
    \caption{Screenshot of the similarity network of an idea under evaluation in OSIP. The thickness of the edge connecting it with other idea and projects is proportional to the value of the similarity between them. Shared concepts are depicted as leaf nodes in the graph. When the user hovers over a project or idea the system displays additional information.}
   \label{fig:osipsimgraph}
\end{figure}

In addition, OSIP evaluators have access to a graph visualization (figure \ref{fig:osipsimgraph}) where they can easily see the most similar projects and ideas to the the idea under evaluation. Such graph is helpful to understand the research context of the idea and how it relates to previously funded research work. 

The idea similarity graph can be used to provide a high level overview of the content of the ideas submitted to the platform. Nodes in the graph are ideas or projects and an edge between two nodes exists when similarity between them has been identified. To identify the clusters of ideas we inject the similarity graph in Gephi\footnote{\url{https://gephi.org}}, a network analysis tool, and apply the Louvain method for community detection~\citep{Blondel_2008}. Once we have the cluster of similar ideas, we aggregate the concepts representing each idea and choose the most frequents as the representative concepts of the clusters. 

\begin{table}[ht!]
\centering
\caption{Clusters extracted from the idea similarity graph in OSIP}
\label{tab:clusters}
\resizebox{0.9\textwidth}{!}{%
\begin{tabular}{p{0.05\textwidth}p{0.95\textwidth}}
\toprule
Ideas & Top Concepts in Cluster                               \\ \midrule
277   & satellite, constellation, orbit,   debris, cost, spacecraft, risk, aim, space debris, European Space Agency                       \\
195   & detection, artificial   intelligence, satellite, European Space Agency, aim, information, plastic, technology,   data, challenge  \\
161   & spacecraft, aim, European   Space Agency, process, information, data, measure, magnetosphere, space   mission, algorithm          \\
156   & rover, exploration, robot, environment,   challenge, cave, space mission, Mars, Moon, project                                     \\
140 & algorithm, machine learning, quantum,   information, data, artificial intelligence, European Space Agency, optimization,   quantum computer, process \\
136   & regolith, construction, habitat,   Moon, process, sintering, technology, European Space Agency, equipment, material               \\
133   & detector, satellite, information,   data, equipment, artificial intelligence, cost, solution, radiation, European   Space Agency  \\
118   & antenna, satellite, radio   frequency, frequency, aim, laser beam, information, equipment, constraint, communications   satellite \\
115   & asteroid, atmosphere, spacecraft,   aim, Venus, European Space Agency, space mission, orbit, process, world                       \\
107 & artificial intelligence, machine   learning, solution, detection, data, learning, process, information, algorithm,   European Space Agency           \\
      \bottomrule
\end{tabular}%
}
\end{table}

The idea clusters we obtained from OSIP are presented in table \ref{tab:clusters}. The risk of debris in the space is a topic that is highly represented in the ideas submitted to the platform. In addition, other topics such as the detection of plastics using satellites and AI, or mars and moon exploration via rovers are also very frequent. OSIP managers can tap into these clusters when planning the future campaigns to gather ideas around subjects that are not well studied, or to avoid funding ideas on topics that are already well covered. More critically some clusters might be seen as a forecast of the topics that could become trendy and important in the future. 


\section{Recommendations and lessons learnt}
\label{sec:disc}

In previous sections, we have seen four case studies of text analytics and natural language processing and understanding in space, ranging across the spectrum of applications of AI and particularly NLP in space. The case study in section~\ref{sec:spaceqa} showcases the innovative application of state-of-the-art deep learning technologies to solve language understanding challenges, like answering questions about space based on a collection of long technical documents. Next, section~\ref{sec:spaceqquiz} focuses on the automatic generation of quizzes with questions about key knowledge related to ESA quality management procedures, illustrating the application of text generation technologies in space. The following case study in section~\ref{sec:fair} applies information extraction to produce machine-readable metadata from space documents and scientific literature in key scientific disciplines to contribute to long-term preservation, making information easier to access by humans and machines. Finally, the case study in section~\ref{sec:osip} deals with determining how innovative an idea submitted to OSIP can be compared to previous ideas, studies, and projects. 

In all such case studies, the guidelines proposed in section~\ref{sec:fw} were considered, i.e. to define each use case, identify the relevant NLP tasks, assess resource availability, and finally decide whether to opt for a machine learning-based approach, a knowledge-based approach or a combination of both. As a general good practice we tried to make results explainable, understood as the degree to which an observer can understand the cause of a model prediction~\citep{Biran2017ExplanationAJ,Buijsman2022DefiningEA}, and to minimize the environmental footprint of our models, e.g. by fine-tuning pre-trained models and applying transfer learning rather than training from scratch. 

We think such case studies constitute a representative sample of NLP projects in space involving either information extraction, comprehension tasks or both. As shown in several of them, for many information management challenges in space like document categorization, search, recommendation, and several forms of information extraction, one of the key components is often a domain-specific knowledge graph that explicitly represents the main concepts and entities of the domain while at the same time describing how such elements are related to each other. The extended knowledge graph for space shown in sections~\ref{sec:fair} and~\ref{sec:osip} is a good example of such kind of structured resources. We recommend to continue dedicating effort to improve such resources, increasing both quality and coverage of the space domain, in addition to other initiatives involving related resources like the ESA Technology Tree and Space Taxonomy. At the same time, in the transit between AI that assists humans to more easily find and access information to AI that is able to understand and reason with scientific documents, we witness an increasing role of deep learning approaches and particularly neural language models. Therefore, the availability of data to fine-tune such models for specific NLP tasks in space is of key importance. Special attention should be given to the production of such datasets, as well as to their distribution and general availability. 

Current methods to train large language models are hardware-intensive, require large amounts of text data to train them, and such training comes at the cost of high energy consumption and a large carbon footprint. Because of this, most of the neural language models available nowadays, like BERT~\citep{devlin-etal-2019-bert}, RoBERTa~\citep{liu2019roberta}, T5~\citep{raffelT52020}, GPT-3~\citep{brown2020language}, etc., have been trained on general-purpose documents collected from the internet and freely available resources, which a priori would hinder their application in vertical domains like space, requiring additional pre-training on domain-specific data that is not easy to find. Few are the examples of neural language models pre-trained on such data, like SpaceRoBERTa, recently released by~\cite{9548078}. However, more research is still needed to show a significant impact on downstream tasks like question answering and question generation in space (see experimental results in sections~\ref{sec:spaceqa} and~\ref{sec:spaceqquiz}) over models trained by fine-tuning general-purpose language models like RoBERTa on general-purpose labeled datasets like SQuAD~\citep{rajpurkar2018know}. 

The type of data required to train text analytics and NLP models can vary according to the task at hand. Downstream tasks such as named entity recognition, summarization, question answering, and question generation typically require training and test data to be labeled. However, data labeling can be a time and effort-intensive task that often requires skilled domain expertise, which can be a costly overhead. The lack of in-house expertise to create labeled datasets has increased the demand for third-party data providers. In addition, online platforms such as Amazon’s Mechanical Turk\footnote{\url{https://www.mturk.com}} are also popular for (trivial, non-expert) labeling tasks. Nevertheless, our experiments show evidence that the impact of applying labeled data to fine-tuning pre-trained language models to solve language understanding tasks like question answering and question generation in space can be greater in practice than investing a similar amount of effort in generating a domain-specific language model on space data. More research is needed to identify the sweet spot in the distribution of effort between language model pre-training and fine-tuning in order to solve domain-specific task in space optimally and maximizing the cost-benefit ratio involved in data management. In any case, we recommend organizations like ESA to continue investing on an internal culture of dataset creation and curation, including the annotation of text corpora for downstream NLP tasks like the ones addressed in this paper. 

\section{Conclusions}
\label{sec:conc}

The European Space Agency (ESA) helps to answer the biggest scientific questions of our time, such as the mysteries of the Universe, the understanding of our Solar System and the quest for habitable planets. The amount, depth and scope of the data, information and knowledge generated and managed throughout the lifecycle of the different space missions is enormous and their contribution to scientific progress is invaluable. 

From the announcement of opportunity and feasibility study to space and ground segment design, development, operations, mission decommissioning and long-term preservation, a large amount of heterogeneous data and information is produced that needs to be managed. The source of ESA information assets is equally diverse and may range from open calls for ideas to develop innovative technology stemming from the wider technical and scientific space community to concurrent design facility documents, technical reports, operation and quality management procedures, and space records relevant to missions spanning over more than 40 years, like climate data records, exploitation reports or scientific publications. 

Managing, mining, and exploiting such wealth of information, of which a large part is free text, addressing silos and ensuring interoperability is a colossal task that goes beyond human capabilities and requires automation. In this paper, we propose a methodological framework for the application of NLP technologies to address such challenges and illustrate it through different case studies with different needs and objectives. The results of our work have already impacted on several areas of ESA including Long-Term Data Preservation (LTDP), ESA Records and Information Management (ESA Archives), the Open Space Innovation Platform (OSIP), the Concurrent Design Facility (CDF), and ESOC Operations Quality Management. We hope that our efforts contribute to the creation of a systematic approach to addressing NLP challenges at ESA, and more generally in space. We also think that part of our  approach can also be extensible to other areas of AI dealing with unstructured data in addition to language, like vision, facilitating their application in space. 

\section*{Acknowledgements}

We are grateful to ESA and the European Commission for the support received to carry out this research. The work presented in sections~\ref{sec:spaceqa} and~\ref{sec:spaceqquiz} was funded by ESA under contract AO/1-10291/20/D/AH - \textit{Text and Data Mining to Support Design, Testing and Operations}. The work presented in section~\ref{sec:fair} was partially funded by ESA as a long-term data preservation activity, with ESRIN Purchase Order 5001024309, and by the EC Horizon2020 project Reliance, under grant 101017501. Finally, the work presented in section~\ref{sec:osip} was funded by the activity \textit{Exploring AI capabilities to support the evaluation process of ideas submitted in the OSIP platform}, with ESA Contract Nr. 4000129447/19/NL/AS.



\bibliographystyle{elsarticle-harv} 
\bibliography{NLPinSpace}





\end{document}